\documentclass[10pt,twocolumn,letterpaper]{article}

\usepackage[pagenumbers]{cvpr}

\usepackage{bm}

\definecolor{Highlight}{HTML}{39b54a}  

\usepackage{amssymb}
\usepackage{pifont}
\usepackage{algorithm}
\usepackage{listings}
\usepackage{etoolbox}
\makeatletter
\AfterEndEnvironment{algorithm}{\let\@algcomment\relax}
\AtEndEnvironment{algorithm}{\kern2pt\hrule\relax\vskip3pt\@algcomment}
\let\@algcomment\relax
\newcommand\algcomment[1]{\def\@algcomment{\footnotesize#1}}
\renewcommand\fs@ruled{\def\@fs@cfont{\bfseries}\let\@fs@capt\floatc@ruled
  \def\@fs@pre{\hrule height.8pt depth0pt \kern2pt}%
  \def\@fs@post{}%
  \def\@fs@mid{\kern2pt\hrule\kern2pt}%
  \let\@fs@iftopcapt\iftrue}
\makeatother

\definecolor{cvprblue}{rgb}{0.21,0.49,0.74}
\usepackage[pagebackref,breaklinks,colorlinks,citecolor=cvprblue]{hyperref}
\usepackage{multirow}
\usepackage{changepage}
\usepackage{mathrsfs}
\usepackage{amsmath}
\usepackage{amssymb}
\usepackage[table,dvipsnames,x11names]{xcolor}
\usepackage{makecell}
\usepackage{tabularx}
\usepackage{float}
\usepackage[accsupp]{axessibility}  

\setitemize{noitemsep,topsep=0pt,parsep=0pt,partopsep=0pt}
\iftoggle{arxiv}{}{
\expandafter\def\expandafter\normalsize\expandafter{%
    \normalsize%
    \setlength\abovedisplayskip{4pt}
    \setlength\belowdisplayskip{4pt} 
    \setlength\abovedisplayshortskip{0pt}%
    \setlength\belowdisplayshortskip{0pt}%
}
}
\newcommand\ourmodel{\emph{Omni-Attribute}\xspace}
\newcommand{\myparagraph}[1]{\noindent\textbf{#1}}
\newcommand{\mycaption}[2]{\caption{\textbf{#1.}~#2}}
\newcommand{\cellfirst}{\cellcolor{Green!50}}
\newcommand{\cellsecond}{\cellcolor{Green!30}}
\newcommand{\cellthird}{\cellcolor{LimeGreen!25}}

\title{\ourmodel: Open-vocabulary Attribute Encoder for \\ Visual Concept Personalization}

\author{
{\large Tsai-Shien Chen$^{1,2,*}$ \: Aliaksandr Siarohin$^1$ \: Gordon Guocheng Qian$^1$ \: Kuan-Chieh Jackson Wang$^1$} \\
{\large Egor Nemchinov$^1$ \: Moayed Haji-Ali$^1$ \:  Riza Alp Guler$^1$ \:  Willi Menapace$^1$ \: Ivan Skorokhodov$^1$} \\
{\large Anil Kag$^1$ \: Jun-Yan Zhu$^{3}$ \: Sergey Tulyakov$^1$} \\
{\normalsize $^1$Snap Inc. \quad $^2$UC Merced \quad $^3$CMU} \\
{\small \url{https://snap-research.github.io/omni-attribute}}
}

\begin{document}

\twocolumn[{
\renewcommand\twocolumn[1][]{#1}
\maketitle
\begin{center}
    \centering
    \captionsetup{type=figure}
    \vspace{-6mm}
    \includegraphics[trim={4mm 0 4mm 0}, width=\textwidth]{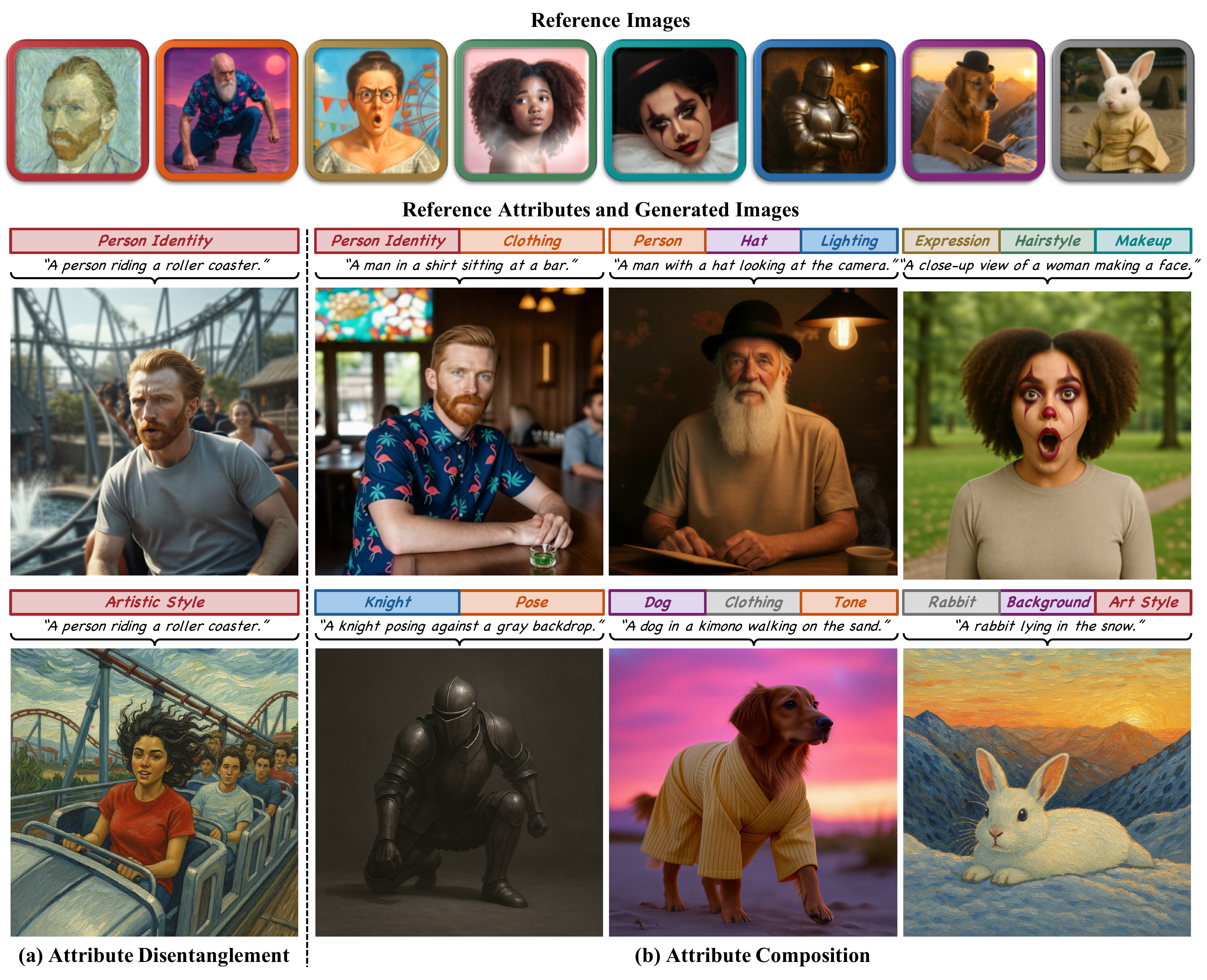}
    \vspace{-7.5mm}
    \caption{
        \ourmodel is an open-vocabulary image attribute encoder that learns to \textit{extract attribute-specific representations} from visual inputs. Given reference images (top row) paired with textual attribute descriptions (colored text boxes), \ourmodel encodes attribute representations that can be coherently synthesized in new contexts (middle and bottom rows) in a fully feed-forward manner, without any test-time optimization. These learned representations possess two key properties: (a) they capture high-fidelity, fine-grained details of the user-specified image attributes while suppressing irrelevant visual information, thereby mitigating ``\textit{copy-and-paste}'' artifacts; and (b) they are composable, enabling the seamless integration of representations from multiple images into a single coherent generated image.       
    }
    \label{fig:teaser}
\end{center}

\vspace{-5.5mm}
\noindent\rule{4cm}{0.4pt}
\vspace{-1.2mm}

{\footnotesize $^*$This work was done while interning at Snap.}
}]
\begin{abstract}

Visual concept personalization aims to transfer only specific image attributes, such as identity, expression, lighting, and style, into unseen contexts. However, existing methods rely on holistic embeddings from general-purpose image encoders, which entangle multiple visual factors and make it difficult to isolate a single attribute. This often leads to information leakage and incoherent synthesis. To address this limitation, we introduce \textnormal{\ourmodel}, the first open-vocabulary image attribute encoder designed to learn high-fidelity, attribute-specific representations. Our approach jointly designs the data and model: (i) we curate semantically linked image pairs annotated with positive and negative attributes to explicitly teach the encoder what to preserve or suppress; and (ii) we adopt a dual-objective training paradigm that balances generative fidelity with contrastive disentanglement. The resulting embeddings prove effective for open-vocabulary attribute retrieval, personalization, and compositional generation, achieving state-of-the-art performance across multiple benchmarks.

\end{abstract}

\begin{figure*}[t]
    \centering
    \includegraphics[width=\linewidth]{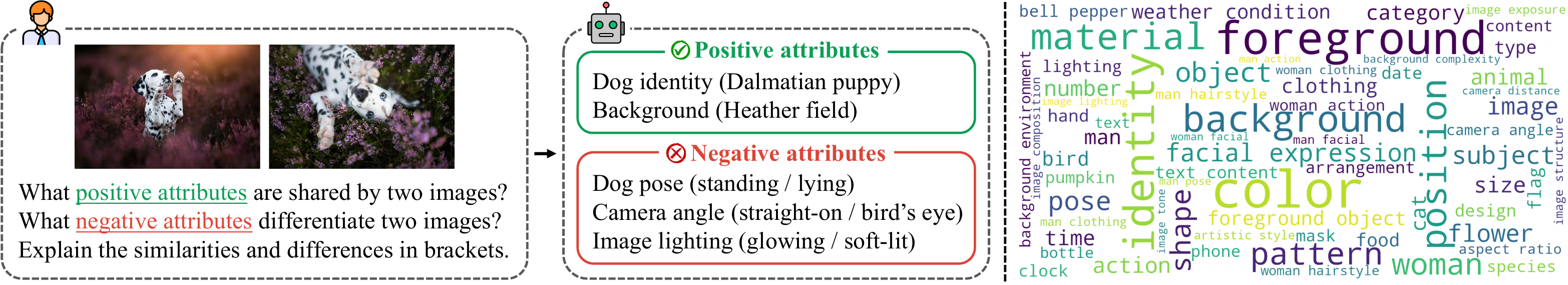}
    \vspace{-4mm}
    \mycaption{Training data annotation}{
        Our training data consist of semantically linked image pairs annotated with positive and negative attributes that define their relationships through the shared and differing characteristics. The word cloud on the right highlights the richness and diversity of our attribute annotations, facilitating the training of an open-vocabulary attribute encoder.
    }
    \label{fig:data_annotation}
\end{figure*}

\section{Introduction}
\label{sec:introduction}

\begin{center}
\vspace{+1.5mm}
\small\emph{“Reality is a mosaic of independent but interconnected things.”} \\
\hfill — Inspired by Gottfried Wilhelm Leibniz
\vspace{+0.5mm}
\end{center}

\noindent Images are bags of visual words~\cite{bag_of_words}, encapsulating rich yet \textit{entangled} combinations of attributes such as identity, expression, pose, background, lighting, camera angle, and art style. This intrinsic complexity makes image attribute \textit{disentanglement} and manipulation a challenging problem. 

Recent advances in personalization~\cite{textual_inversion,ip_adapter} have demonstrated remarkable capabilities in transferring image attributes to novel contexts, enabling applications like ``\textit{generating my dog based on the reference image.}'' To achieve these objectives, most existing methods rely on general-purpose image encoders, such as CLIP~\cite{clip}, DINOv2~\cite{dinov2}, or the VAEs~\cite{vae}, to extract holistic representations of input images that are then used to guide the synthesis. However, this design presents a fundamental limitation. Since these encoders compress and entangle all visual information into a single representation, they often suffer from \textit{information leakage} of irrelevant attributes, resulting in undesirable ``\textit{copy-and-paste}'' artifacts~\cite{video_alchemist}. For example, as shown in the leftmost column of \cref{fig:qualitative_comparison}, even when personalizing the \textit{identity}, existing image embeddings inadvertently transfer the \textit{lighting} and \textit{clothing} details from the reference image.

In this paper, we revisit the attribute manipulation problem from a new perspective and focus on learning attribute-level representations directly on the encoder side. We introduce \ourmodel, the first open-vocabulary image attribute encoder designed to process an image alongside textual attribute descriptions jointly. Unlike general-purpose encoders that indiscriminately capture all visual content, \ourmodel is built with two goals: $(i)$ to extract attribute-related information faithfully and exclusively, and $(ii)$ to suppress other irrelevant visual information.

We jointly design the data and model for such attribute-level representation learning. On the data side, our training samples consist of pairs of semantically linked images. To guide the encoder in learning attribute-specific representations, we introduce a novel annotation form to establish semantic connections between image pairs. It includes $(i)$ \textit{positive attributes} that describe the semantics shared by the two images and $(ii)$ \textit{negative attributes} that highlight the characteristics that differ between them. This pairing structure explicitly teaches the encoder which visual concepts should be preserved and which should be suppressed.

On the model side, we formulate attribute-level representation learning as a dual-objective optimization problem. On one hand, the attribute embeddings are required to capture sufficient information for the high-fidelity reconstruction of the target attributes; on the other hand, they need to discard irrelevant cues from unrelated attributes. To achieve both goals, we minimize two complementary losses: $(i)$ a \textit{generative loss} that ensures embeddings extracted from a reference image can effectively reconstruct its paired image, and $(ii)$ a \textit{contrastive loss} that introduces a repulsive force between embeddings associated with negative attributes. As illustrated in~\cref{fig:teaser}(a), these two losses together drive the encoder to accurately and exclusively disentangle attribute-specific information.

Finally, we demonstrate the versatility of \ourmodel in the task of attribute composition, where multiple attribute embeddings extracted from different images can be seamlessly combined into a single image, as shown in~\cref{fig:teaser}(b). Our key contributions can be summarized as follows:
\begin{itemize}[leftmargin=8pt]
    \item We present \ourmodel, the first open-vocabulary attribute encoder designed to jointly process an image along with a textual attribute description to extract attribute-specific representations.
    \item To learn such attribute-level embeddings, we introduce a new data annotation strategy together with a novel training scheme that balances high-fidelity encoding with the effective suppression of irrelevant information.
    \item We showcase the versatility of \ourmodel across several downstream tasks, including attribute-oriented image retrieval, personalization, and composition. We further visualize its embedding spaces for better interpretability.
\end{itemize}
\section{Related Work}
\label{sec:related_work}

\myparagraph{Visual Representation Learning.}
\ourmodel learns attribute-level embeddings by combining \textit{supervised}, \textit{contrastive}, and \textit{multimodal} learning, which are the three key pillars of visual representation learning over the past few decades. Early approaches, such as AlexNet~\cite{alexnet} and ResNet~\cite{resnet}, relied on \textit{supervised} pretraining on large-scale datasets like ImageNet~\cite{imagenet} to produce hierarchical features transferable across recognition tasks. Subsequently, self-supervised methods~\cite{simclr,moco,byol,swav,ifnd} introduced instance-level \textit{contrastive} objectives to eliminate dependence on labeled data while retaining discriminative representations.

More recently, CLIP~\cite{clip} unified vision and language through \textit{multimodal} representation learning, aligning the embedding spaces of both modalities. Building on this foundation, subsequent works like DINO~\cite{dino,dinov2,dinov3} and MAE~\cite{mae} advanced visual abstraction but still encoded holistic global features that entangle diverse image attributes. Extending this line of work, \ourmodel explicitly models attribute-level representations, producing disentangled and composable embeddings that bridge representation learning and controllable image generation.

\myparagraph{Image-guided Generation.}
The goal of image-guided generation is to manipulate or extend the visual attributes of a reference image while synthesizing the remaining context in a coherent and semantically consistent setting. Recent breakthroughs in diffusion models~\cite{diffusion,ddpm,ddim} and transformer-based architectures~\cite{transformer,vit,snapvideo} have significantly enhanced generation quality, enabling applications such as editing~\cite{sdedit,prompt_to_prompt,plug_and_play,omnigen,flux_kontext,qwen_image_edit,nano_banana} and personalization~\cite{textual_inversion,custom_diffusion,dreambooth,ip_adapter,instantbooth,panda,mcdiff,vimi,video_alchemist,alchemint,layercomposer,canvas2image}.

A common strategy is the encoder-based approach, where a reference image is first mapped to latent embeddings, which are then used to condition the generative model. For instance, IP-adapter~\cite{ip_adapter} injects CLIP-encoded image features via lightweight decoupled cross-attention layers for personalization. Qwen-Image-Edit~\cite{qwen_image_edit} jointly encodes textual and visual instructions using a multimodal encoder~\cite{qwenvl2} and a VAE~\cite{vae} for unified vision–language conditioning. However, these conditioning embeddings often entangle multiple visual attributes, leading to information leakage and undesirable ``\textit{copy-and-paste}'' artifacts~\cite{video_alchemist}. \ourmodel addresses this issue on the encoder side by learning attribute-specific embeddings, resulting in cleaner, more controllable synthesis.

\myparagraph{Visual Concept Disentanglement.}
Images inherently blend multiple visual attributes across shared pixels, making attribute disentanglement a long-standing challenge. Early works, such as Break-A-Scene~\cite{break_a_scene} and ConceptExpress~\cite{conceptexpress}, attempt to separate concepts using user-defined or attention-derived spatial masks, but these methods are limited to isolating spatially separable elements. Inspiration Tree~\cite{inspiration_tree} introduces a hierarchical decomposition of visual concepts but lacks predictability in its representations.

Recent methods like Token-Verse~\cite{tokenverse} (optimization-based) and Mod-Adapter~\cite{mod_adapter} (encoder-based) manipulate the modulation space of DiTs~\cite{dit} to represent image attributes but face two limitations: $(i)$ per-token (word-level) modulation hinders the personalization of multi-token (phrase-level) concepts; and $(ii)$ the usage of AdaLN conditioning restricts control to simple, limited affine transformations (scale-and-shift). Closer to our setting, OADis~\cite{OADis} and DeCLIP~\cite{declip} leverage text-guided contrastive objectives for attribute disentanglement, but they are restricted to a fixed, closed set of attributes. In contrast, \ourmodel is able to extract open-vocabulary attribute embeddings, enabling more flexible image generation.
\section{\ourmodel}
\label{sec:methodology}

Our target is to learn an open-vocabulary attribute encoder that jointly takes images and textual attribute descriptions as inputs and produces disentangled, attribute-specific representations while suppressing other visual information.

\subsection{Semantic Connections between Image Pairs}
\label{sec:methodology_annotation}

As illustrated in~\cref{fig:data_annotation}, our training samples consist of semantically linked image pairs. To learn attribute-level representations, we design a new annotation scheme that establishes semantic connections between each image pair through two types of attributes: $(i)$ \textit{positive attributes}, which describe shared semantic properties, and $(ii)$ \textit{negative attributes}, which highlight distinct characteristics.

\myparagraph{Annotation of Positive and Negative Attributes.}
Labeling high-quality attribute annotations requires a strong vision-language understanding and a long, detailed instruction prompt, which can be prohibitively expensive for large-scale inference. To balance annotation quality and cost, we adopt a two-stage annotation pipeline. In the first stage, we leverage the powerful but computationally expensive 72B-parameter multimodal large-language model (MLLM)~\cite{qwenvl2} with the input of a detailed instruction prompt to curate a sub-dataset with high-quality attribute annotations. Inspired by Chain-of-Thought~\cite{chain_of_thought}, we enhance label quality by explicitly prompting the model to describe fine-grained similarities and differences for each positive and negative attribute (see descriptions in the brackets of~\cref{fig:data_annotation}).

In the second stage, we finetune a 32B-parameter MLLM on these annotated image pairs to learn an expert annotator model specialized in this task. Through supervised finetuning, this student model internalizes the task-specific reasoning behavior and the structured annotation output, enabling it to perform high-quality labeling without the detailed instruction prompt. This substantially reduces annotation costs by cutting input token length by $3.1\times$ and per-sample annotation latency by $6.3\times$. \cref{app:implementation_annotation} details the MLLM finetuning and the annotation pipeline.

The word cloud in~\cref{fig:data_annotation} (right) illustrates the richness and diversity of the resulting attribute annotations, facilitating the learning of an open-vocabulary attribute encoder.

\begin{figure}[t]
    \centering
    \includegraphics[width=\linewidth]{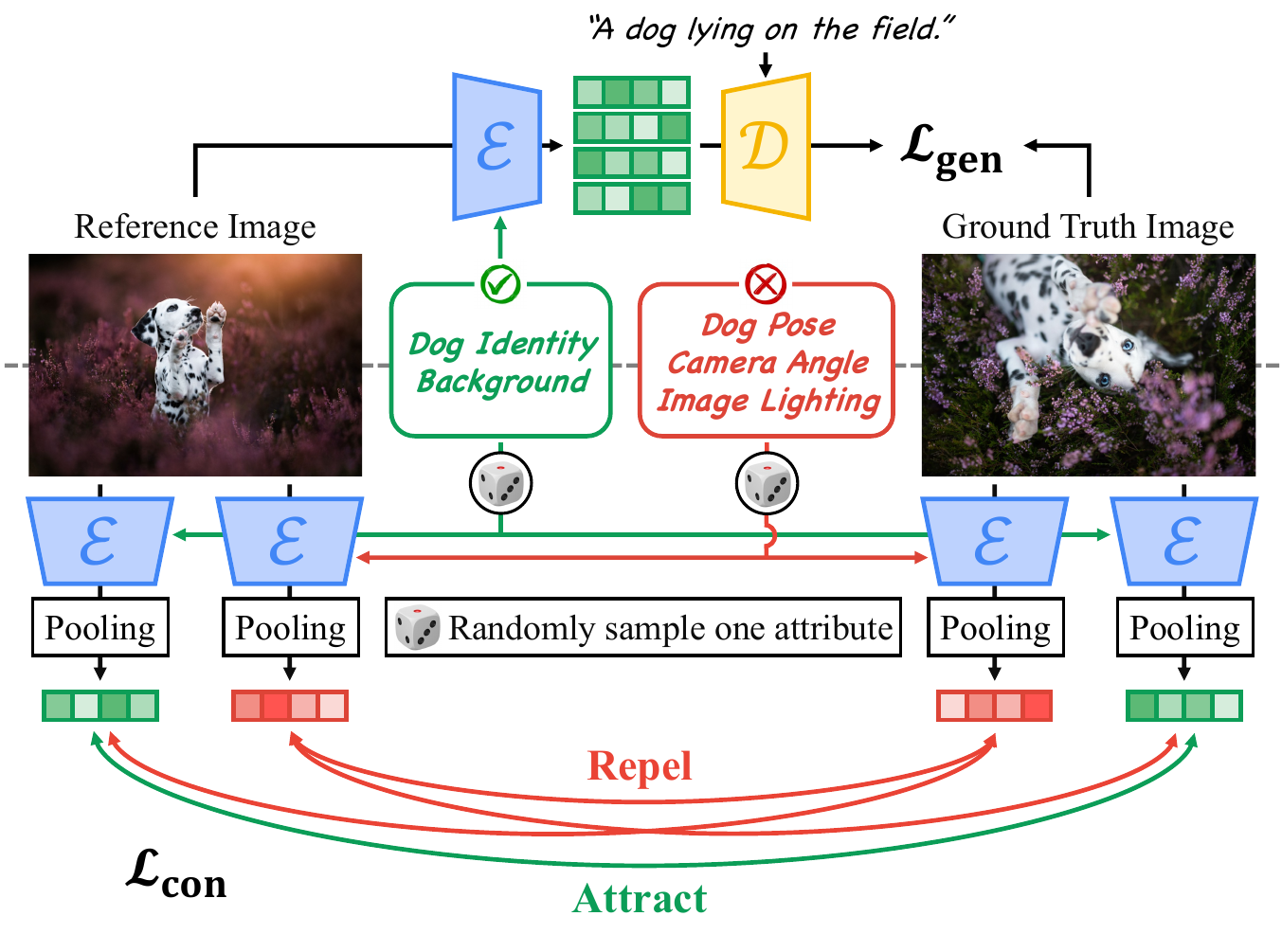}
    \vspace{-6mm}
    \mycaption{Dual-objective representation learning}{
         Our attribute-level representation learning is guided by two complementary objectives: a \textit{generative loss} (top) to \textit{maximize} embedding information and encourage the capture of fine-grained, high-fidelity details, and a \textit{contrastive loss} (bottom) that extracts attribute-specific information while \textit{suppressing} irrelevant content.
    }
    \label{fig:training_loss}
\end{figure}
\begin{figure}[t]
    \centering
    \includegraphics[width=\linewidth]{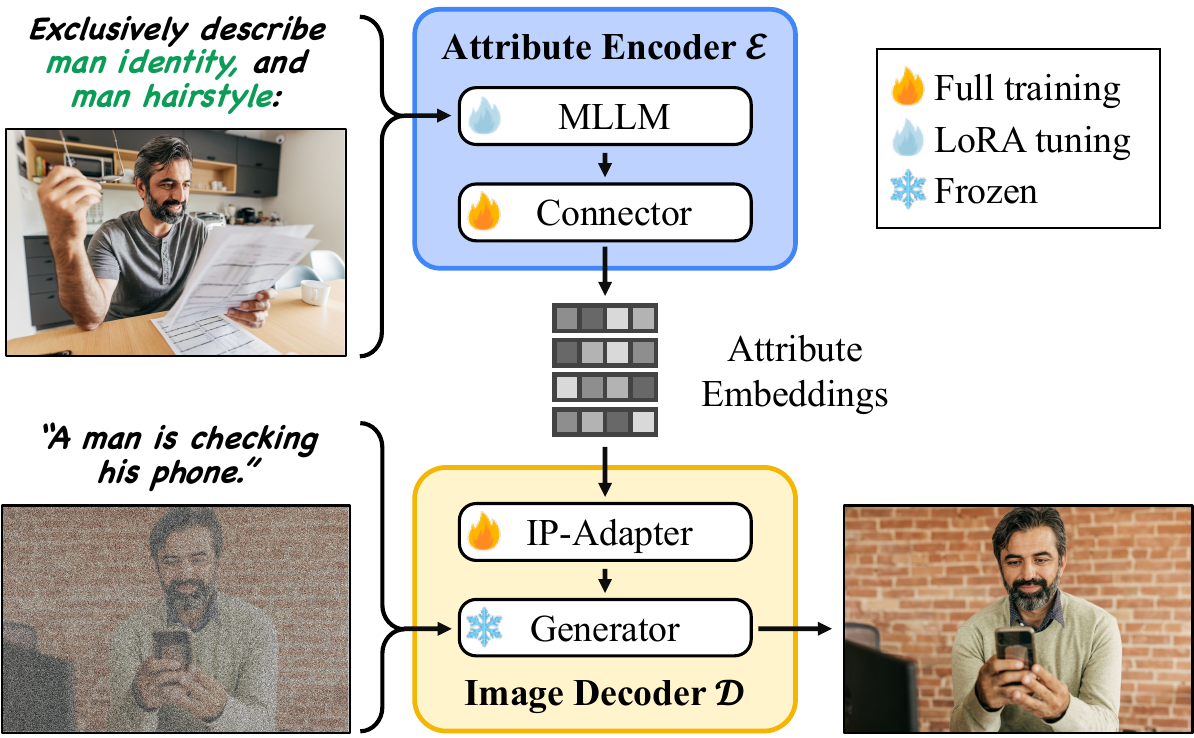}
    \vspace{-3mm}
    \mycaption{Model architecture}{
         Our attribute encoder is a LoRA-tuned MLLM followed by a trainable lightweight connector to preserve strong vision-language prior while capable of adapting to our attribute disentanglement task. The image decoder is a frozen generator with trainable IP-Adapter~\cite{ip_adapter} modules for personalization.
    }
    \label{fig:model_architecture}
\end{figure}

\subsection{Attribute-level Representation Learning}

Learning high-fidelity, attribute-level embeddings is inherently a dual-objective optimization problem. On one hand, the embeddings must \textit{maximize} attribute-specific information to represent fine-grained details. On the other hand, they must \textit{suppress} signals from irrelevant attributes. As in~\cref{fig:training_loss}, we jointly optimize two complementary losses to achieve both goals: $(i)$ a \textit{generative loss} that encourages faithful attribute reconstruction and $(ii)$ a \textit{contrastive loss} that enforces attribute separation and disentanglement.

\myparagraph{Generative Loss.}
The training framework consists of an encoder $\mathcal{E}$ and a decoder (or generator) $\mathcal{D}$. Given a training image pair $(I_x, I_y)$ with $m$ positive attributes $\{a^+_1, \ldots, a^+_m\}$ and $n$ negative attributes $\{a^-_1, \ldots, a^-_n\}$, we randomly assign one image as the reference image $I_r$ and the other as the ground truth image $I_g$ to compute the generative loss. As shown in~\cref{fig:training_loss} (top), the generative loss guides the model to reconstruct $I_g$ conditioned on the attribute embeddings extracted from $I_r$ and its corresponding text prompt $c_g$:
\begin{equation}
\label{eq:generative}
    \mathcal{L}_{\mathrm{gen}} = \phi({I^*}, I_g), \hspace{0.4em} I^* = \mathcal{D}(\mathcal{E}(I_r, \{a^+_1, \ldots, a^+_m\}), c_g)
\end{equation}
where $\phi$ is a generic similarity or distance function between images (\eg, $\mathrm{L_2}$ or flow-matching loss). Notably, \textit{all} positive attributes are used as inputs here to guide $\mathcal{E}$ in extracting complete attribute information shared by both images. Empirically, dropping any positive attribute during training causes $\mathcal{E}$ to encode the entire image and further leads to the ``\textit{copy-and-paste}'' artifacts~\cite{video_alchemist} rather than focusing on the specified attributes.

\myparagraph{Contrastive Loss.}
To enhance attribute disentanglement, we introduce a contrastive loss that attracts the positive attribute embeddings encoded from the paired images $(I_x, I_y)$, while repelling embeddings associated with negative or different attributes, as depicted in~\cref{fig:training_loss} (bottom). Formally, we sample one positive attribute $a^+_i$ and one negative attribute $a^-_j$, and optimize:
\begin{equation}
\label{eq:contrastive}
    \mathcal{L}_{\mathrm{con}} = -\log\frac{\scriptstyle \psi(a^+_i, a^+_i)}{\scriptstyle \psi(a^+_i, a^+_i) + \psi(a^+_i, a^-_j) + \psi(a^-_j, a^+_i) + \psi(a^-_j, a^-_j)}
\end{equation}
The similarity function $\psi$ is defined as:
\begin{equation}
\label{eq:similarity}
    \psi(a_x, a_y) = \mathrm{sim}(\mathrm{pool}(\mathcal{E}(I_x, a_x)), \mathrm{pool}(\mathcal{E}(I_y, a_y))),
\end{equation}
where $\mathrm{sim}(u, v) = \exp(\frac{1}{\tau}\frac{(u \cdot v)}{\|{u}\|\|{v}\|})$ measures the similarity of two pooled attribute embeddings, with a temperature of $\tau$. This loss encourages the encoder to maximize the similarity between the embeddings of the positive attribute while minimizing the similarity across the negative or different attributes, even when the embedding pairs are always derived from the same image pair $(I_x, I_y)$. Thereby, it produces a discriminative attribute-level embedding space.

The final training objective combines both generative and contrastive losses:
\begin{equation}
\label{eq:total_loss}
    \mathcal{L} = \lambda_{\mathrm{gen}} \cdot \mathcal{L}_{\mathrm{gen}} + \lambda_{\mathrm{con}} \cdot \mathcal{L}_{\mathrm{con}}
\end{equation}
with hyperparameters $\lambda_{\mathrm{gen}}$ and $\lambda_{\mathrm{con}}$ balancing reconstruction fidelity against attribute disentanglement.

\subsection{Model Architecture}
\label{sec:methodology_architecture}

\cref{fig:model_architecture} illustrates our model architecture, which consists of an attribute encoder $\mathcal{E}$ and an image decoder $\mathcal{D}$.

\myparagraph{Attribute Encoder.}
The encoder is designed to satisfy two key requirements: $(i)$ the ability to jointly process text and image inputs, and $(ii)$ a strong vision-language prior to support our attribute disentanglement objective. To meet these criteria, we adopt a MLLM~\cite{qwenvl2} as the backbone. Empirically, we observe that LoRA~\cite{lora} tuning better preserves pretrained representations and mitigates catastrophic forgetting compared to full finetuning, consistent with the recent findings of Shuttleworth~\etal~\cite{lora_vs_full}. To further adapt the model to the attribute disentanglement task, we attach a lightweight, trainable connector~\cite{step1x_edit} after the backbone. 

The encoder takes as input a multimodal prompt composed of the attribute descriptions and an image, as shown in \cref{fig:model_architecture} (upper-left), and produces a sequence of $l$ tokens, $\boldsymbol{A} = [\boldsymbol{a}_1, \ldots, \boldsymbol{a}_l]$, serving as attribute embeddings.

\myparagraph{Contrastive Head.}
To compute the contrastive loss, we aggregate the 2-D attribute embeddings into a 1-D representation via average pooling: $\mathrm{pool}(\boldsymbol{A}) = \sum_{i=1}^{l} \boldsymbol{a}_i/l$.

\myparagraph{Image Decoder.}
For the generative loss, the full attribute embeddings $\boldsymbol{A}$ are directly passed into a downstream decoder, which consists of a frozen image generator~\cite{flux} preceded by trainable IP-Adapter~\cite{ip_adapter} for personalization.

\subsection{Composition of Multiple Attributes}
\label{sec:methodology_composition}

Compositional image generation aims to synthesize a coherent output by integrating multiple objects or concepts from different reference sources. We empirically find that the learned attribute embeddings are \textit{composable}, allowing \ourmodel to achieve compositional generation.

Composable Diffusion~\cite{composable} achieves multi-condition synthesis by generalizing classifier-free guidance~\cite{cfg} (CFG) to handle multiple conditioning signals. Specifically, it combines ``\textit{conditional score directions},'' each of which represents the gradient direction associated with a distinct concept and is obtained by taking the difference between the model's conditional and unconditional predictions.

We adopt a similar concept in our flow-matching~\cite{flow_matching} generator. Given a reference image set $\{I_1, \ldots, I_N\}$ with associated attribute descriptions $\{a_1, \ldots, a_N\}$, where $N$ denotes the number of reference sources, we first compute the \textit{conditional flow field} for each image-attribute pair:
\begin{equation}
\label{eq:single_cfg}
    \Delta_{(I_i, a_i)} = \mathcal{D}(\mathcal{E}(I_i, a_i), \varnothing) - \mathcal{D}(\varnothing, \varnothing).
\end{equation}
We then evaluate the final velocity based on the linear combination of the \textit{conditional flow fields}:
\begin{equation}
\label{eq:multi_cfg}
    v^* = \mathcal{D}(\varnothing, c) + \sum_{i=1}^{N} w_i \cdot \Delta_{(I_i, a_i)},
\end{equation}
where $c$ is the prompt and $w_i$ controls the strength of each conditioning signal. Note that~\cref{eq:multi_cfg} only applies CFG to the attribute conditions. In practice, we also apply CFG on $c$ following the formulation of InstructPix2Pix~\cite{instructpix2pix}.
\section{Experiments}
\label{sec:experiments}

In this section, we demonstrate the versatility of \ourmodel across several downstream tasks, including attribute-oriented retrieval, personalization, and composition. Implementation details are provided in~\cref{app:implementation}.

\begin{figure*}[t]
    \centering
    \includegraphics[width=\linewidth]{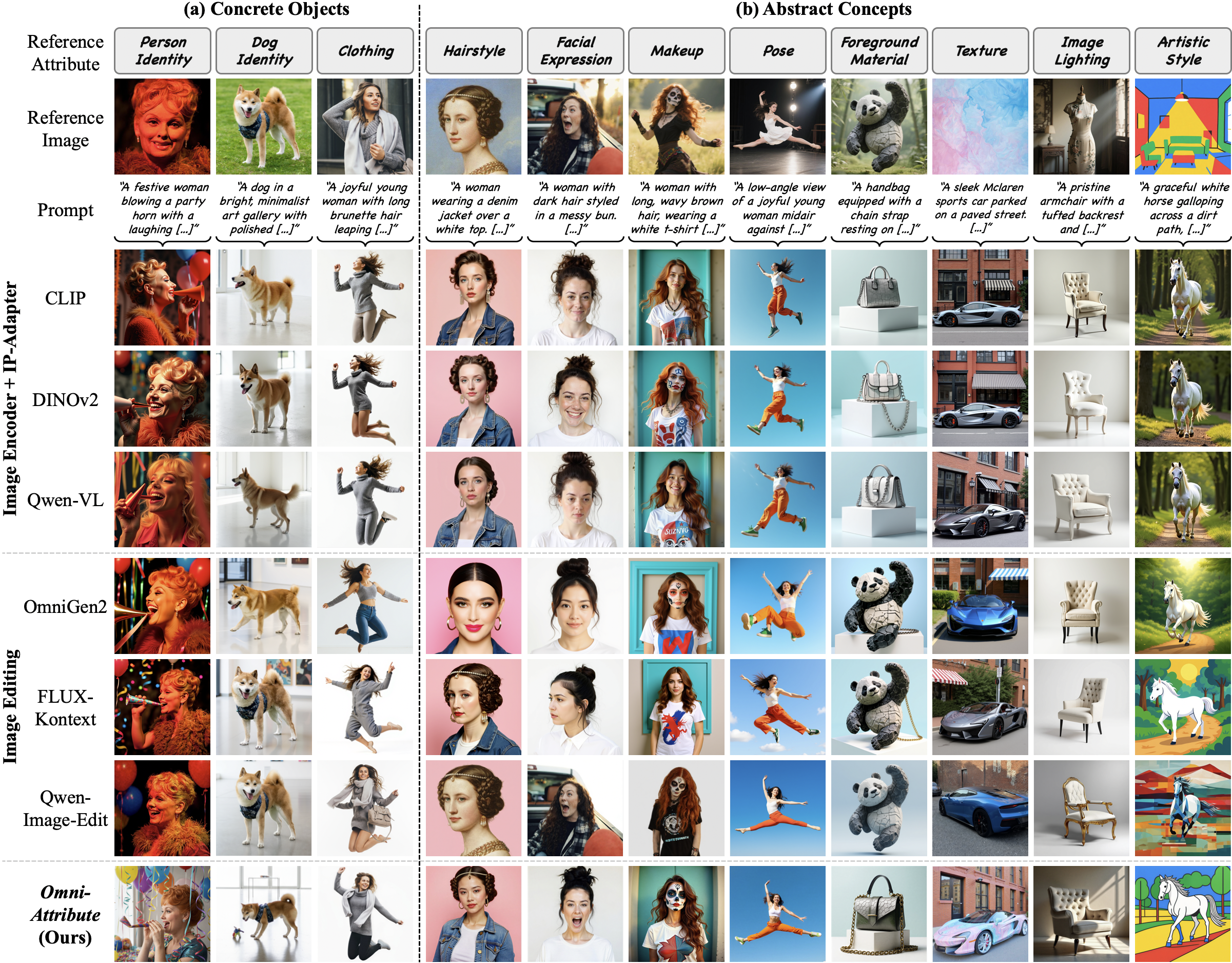}
    \vspace{-5mm}
    \mycaption{Qualitative comparisons of open-vocabulary attribute personalization}{
         Each row, from top to bottom, shows $(i)$ the reference image-attribute pair and the prompt, $(ii)$ results generated using CLIP~\cite{clip}, DINOv2~\cite{dinov2}, and Qwen-VL~\cite{qwenvl2} embeddings, $(iii)$ results from editing models, including OmniGen2~\cite{omnigen}, FLUX-Kontext~\cite{flux_kontext}, and Qwen-Image-Edit~\cite{qwen_image_edit}, and $(iv)$ results by \ourmodel. As shown, \ourmodel achieves the best balance between faithfully encoding the target attribute and coherently synthesizing it into new contexts aligned with the prompt, while minimizing undesired ``\textit{copy-and-paste}'' artifacts~\cite{video_alchemist}. Full prompts are provided in~\cref{app:evaluation_personalization}.
    }
    \label{fig:qualitative_comparison}
\end{figure*}
\begin{figure*}[t]
    \centering
    \includegraphics[width=\linewidth]{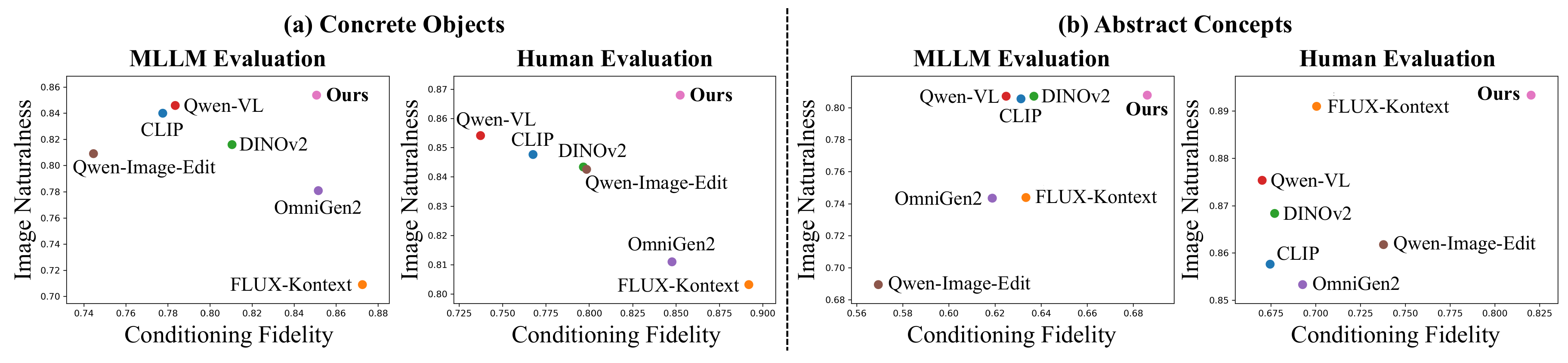}
    \vspace{-6mm}
    \mycaption{Quantitative comparisons of open-vocabulary attribute personalization}{
         We compare \ourmodel with baseline methods on the personalization of two types of attributes: (a) \textit{concrete objects} and (b) \textit{abstract concepts}. We perform the evaluation across two metrics, image naturalness (higher is better) and conditioning fidelity (higher is better), using both MLLM~\cite{gpt_4o} and human evaluations. \ourmodel consistently outperforms existing methods, especially for abstract concepts. Full numerical results are in~\cref{app:evaluation_personalization}.
    }
    \label{fig:quantitative_comparison}
\end{figure*}
\begin{figure}[t]
    \centering
    \includegraphics[width=\linewidth]{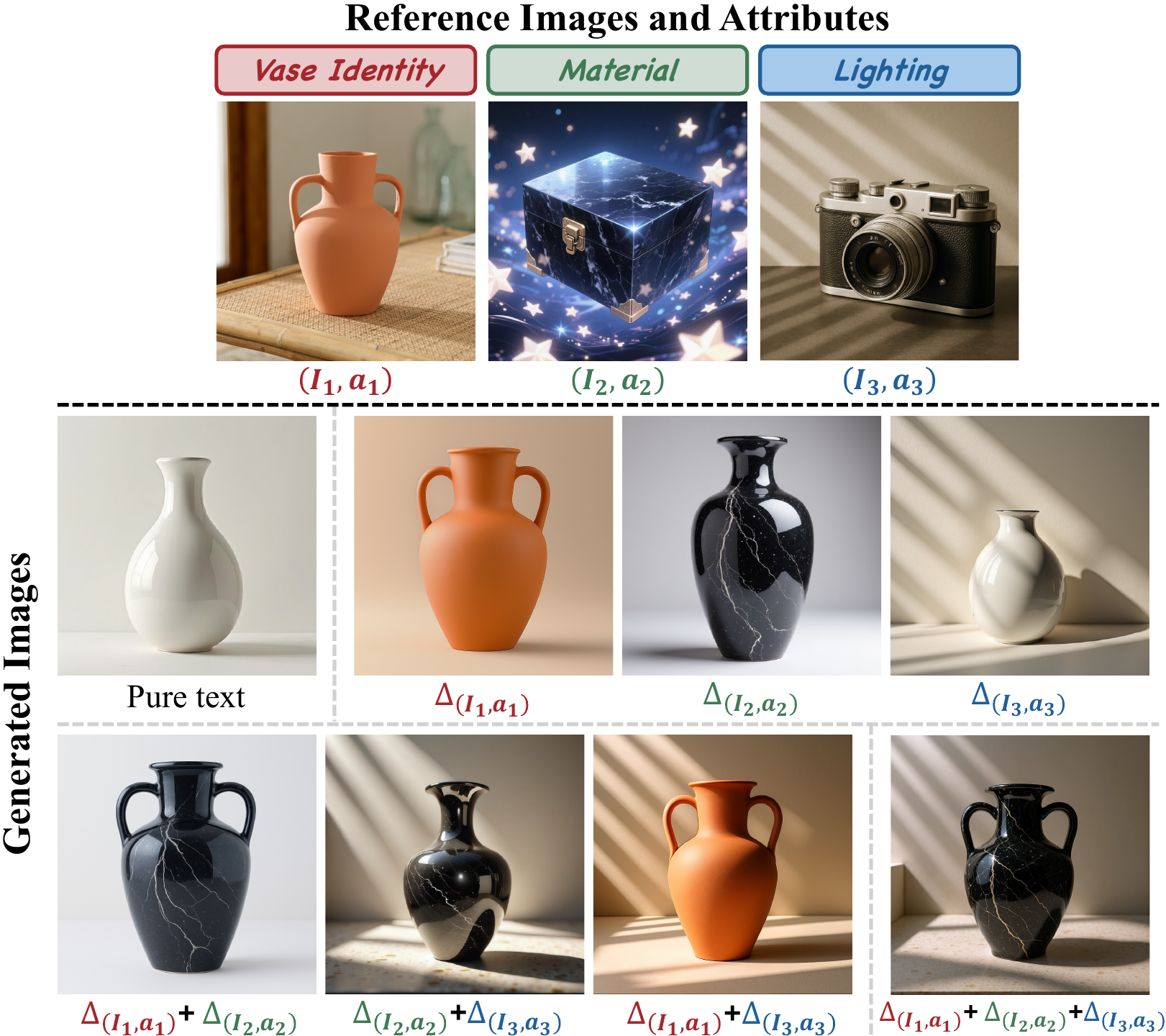}
    \vspace{-3mm}
    \mycaption{Composability of attribute embeddings}{
        From top to bottom, each row shows the input conditions, the effect of a single image-attribute pair, and the compositional results of multiple attributes, showing the composability of our attribute embeddings. The prompt is ``\textit{A vase is standing against a plain background.}''
    }
    \label{fig:attribute_composability}
\end{figure}

\subsection{Open-vocabulary Attribute Personalization}
\label{sec:experiments_personalization}

We compare \ourmodel with existing methods for open-vocabulary attribute personalization and present qualitative and quantitative results in \cref{fig:qualitative_comparison,fig:quantitative_comparison}, respectively.

\myparagraph{Baseline Models.}
We conduct extensive comparisons with two groups of models that support image personalization:

First, we evaluate different image encoders, including CLIP~\cite{clip}, DINOv2~\cite{dinov2}, and Qwen-VL~\cite{qwenvl2}, by training IP-Adapter modules between each encoder and the same frozen image generation backbone to ensure a fair comparison. For Qwen-VL, we provide a multimodal prompt that includes both the reference attribute and image, as shown in~\cref{fig:model_architecture} (upper-left). For vision-only encoders, such as CLIP and DINOv2, we only provide the reference image.

Next, recent advances in image editing enable personalization through models such as OmniGen2~\cite{omnigen}, FLUX-Kontext~\cite{flux_kontext}, and Qwen-Image-Edit~\cite{qwen_image_edit}. To adapt these models for attribute personalization, we reformulate the instruction prompt as: \textit{“Preserve the $<$\underline{attribute}$>$ of the image and generate $<$\underline{prompt}$>$.”}

\myparagraph{Evaluation Dataset.}
To comprehensively evaluate open-vocabulary attribute personalization, we construct a benchmark comprising 15 reference attributes across two categories: \textit{concrete objects} and \textit{abstract concepts} (see the full list in~\cref{app:evaluation_personalization}). For each attribute, we select 5 images (partially sourced from DreamOmni2~\cite{dreamomni2}) and apply an LLM to generate 5 prompts that deliberately exclude content related to the reference attribute to avoid semantic conflicts. By cross-pairing the images and prompts, we obtain 25 samples per attribute, resulting in a total of 375 samples.

\myparagraph{Evaluation Metrics.}
Personalization aims to faithfully integrate the reference visual concept into new contexts specified by a text prompt while maintaining a coherent overall composition. We evaluate this objective using three metrics: $(i)$ \textit{attribute fidelity score}, which measures the faithfulness of the personalized attributes; $(ii)$ \textit{text fidelity score}, which assesses the semantic consistency between the generated image and the input prompt; and $(iii)$ \textit{image naturalness score}, which evaluates the overall visual coherence of the image. An effective personalization approach should achieve balanced performance across all three metrics. Following DreamBench++~\cite{dreambench++}, we employ GPT-4o~\cite{gpt_4o} to assign scores to each generated image on a scale from 0 (poor) to 10 (excellent). We normalize the results to the scale of [0, 1] and report them in~\cref{fig:quantitative_comparison}, where the attribute fidelity and text fidelity are averaged into a single conditioning fidelity score for visualization purposes. \cref{app:evaluation_personalization} includes the original numerical results.

\myparagraph{User Study.}
To complement automated evaluation, we conduct a user study with 10 participants. Each participant rates the three metrics described above, resulting in 11.25K individual ratings. Details of the user study are provided in~\cref{app:evaluation_personalization}, and the results are shown in~\cref{fig:quantitative_comparison}.

\cref{fig:qualitative_comparison} illustrates that CLIP~\cite{clip} and DINOv2~\cite{dinov2} struggle to personalize \textit{abstract concepts} due to the lack of support from reference attribute inputs. Although Qwen-VL~\cite{qwenvl2} accepts additional attribute inputs, the encoder lacks architectural refinement and attribute-level contrastive learning, which results in poor adaptation to attribute personalization. In contrast, image editing models~\cite{omnigen,flux_kontext,qwen_image_edit} can generate outputs that more closely resemble the reference image but often fail to disentangle the target attribute, causing noticable ``\textit{copy-and-paste}'' artifacts~\cite{video_alchemist} and weak text alignment. By comparison, \ourmodel achieves the best balance between image naturalness and alignment with both text and attribute conditioning. These observations are consistent with the quantitative results from both MLLM and human evaluations presented in~\cref{fig:quantitative_comparison}.

\subsection{Compositional Image Generation}
\label{sec:experiments_composition}

As described in~\cref{sec:methodology_composition}, we empirically find that our attribute embeddings can achieve compositional image generation through the linear combination of ``\textit{conditional flow fields},'' as defined in~\cref{eq:single_cfg,eq:multi_cfg}. Here, we qualitatively demonstrate the effect of each conditional flow field in~\cref{fig:attribute_composability} (middle-right) and then progressively compose them into a single coherent image in~\cref{fig:attribute_composability} (bottom), demonstrating the composability of our attribute embeddings. More results can be found in~\cref{fig:teaser}(b) and~\cref{app:additional_results}.

\begin{table*}[t]
    \centering
    \small
    \mycaption{Ablation study}{
        From top to bottom, we ablate: [a-b] the connectors with varying numbers of self-attention and linear layers; [b-d] different training strategies for the MLLM (frozen, LoRA, or full finetuning); [e-h] the hyperparameters of the contrastive loss; and [i] our final setting. We report two sets of results: $(i)$ the gap between the cosine similarities of positive and negative attributes and $(ii)$ text fidelity, attribute fidelity, image naturalness, and their average on attribute personalization. We color the models with top 3 performance in the groups of [a-d] and [e-h], respectively.
    }
    \label{tab:ablation_study}
    \begin{tabular}{l|cc|cc|c|cccc}
        \toprule
        \multirow{2}{*}{Index} &
        \multicolumn{2}{c|}{Attribute Encoder} &
        \multicolumn{2}{c|}{Contrastive Loss} &
        \multicolumn{1}{c|}{Cosine Sim.} &
        \multicolumn{4}{c}{Attribute Personalization} \\
        & MLLM & Connector & Weight $\lambda_{\mathrm{con}}$ & Temp. $\tau$ & $\Delta_{(\textrm{pos}, \textrm{neg})}$$\uparrow$ & Text-F$\uparrow$ & Attr-F$\uparrow$ & Natural$\uparrow$ & Average$\uparrow$ \\
        \midrule
        \multicolumn{10}{c}{\textit{Ablation of Attribute Encoder Design and Training}} \\
        \midrule
        {}[a] & Frozen & 1 Linear               & 0    & -     & 0.003  & 0.929 & 0.479 & 0.827 & 0.745 \\
        {}[b] & Frozen & 8 Self-Attn + 1 Linear & 0    & -     & 0.003  & 0.920 & 0.494 & 0.823 & \cellthird 0.746 \\
        {}[c] & LoRA   & 8 Self-Attn + 1 Linear & 0    & -     & -0.002 & 0.873 & 0.651 & 0.797 & \cellfirst 0.774 \\
        {}[d] & Full   & 8 Self-Attn + 1 Linear & 0    & -     & -0.003 & 0.867 & 0.600 & 0.774 & \cellsecond 0.747 \\
        \midrule
        \multicolumn{10}{c}{\textit{Ablation of Hyperparameters of Contrastive Loss}} \\
        \midrule
        {}[e] & LoRA   & 8 Self-Attn + 1 Linear & 0.01  & 0.5  & \cellfirst 0.738  & 0.914 & 0.513 & 0.855 & 0.761 \\
        {}[f] & LoRA   & 8 Self-Attn + 1 Linear & 0.01  & 0.02 & 0.121  & 0.875 & 0.639 & 0.803 & \cellthird 0.772 \\
        {}[g] & LoRA   & 8 Self-Attn + 1 Linear & 0.1   & 0.1  & \cellsecond 0.641  & 0.900 & 0.577 & 0.842 & \cellsecond 0.773 \\
        {}[h] & LoRA   & 8 Self-Attn + 1 Linear & 0.001 & 0.1  & \cellthird 0.502  & 0.883 & 0.640 & 0.812 & \cellfirst 0.778 \\
        \midrule
        {}[i] & LoRA   & 8 Self-Attn + 1 Linear & 0.01  & 0.1  & 0.608  & 0.896 & 0.641 & 0.831 & 0.789 \\
        \bottomrule
    \end{tabular}
\end{table*}
\begin{figure*}[!ht]
    \centering
    \vspace{+3mm}
    \includegraphics[width=\linewidth]{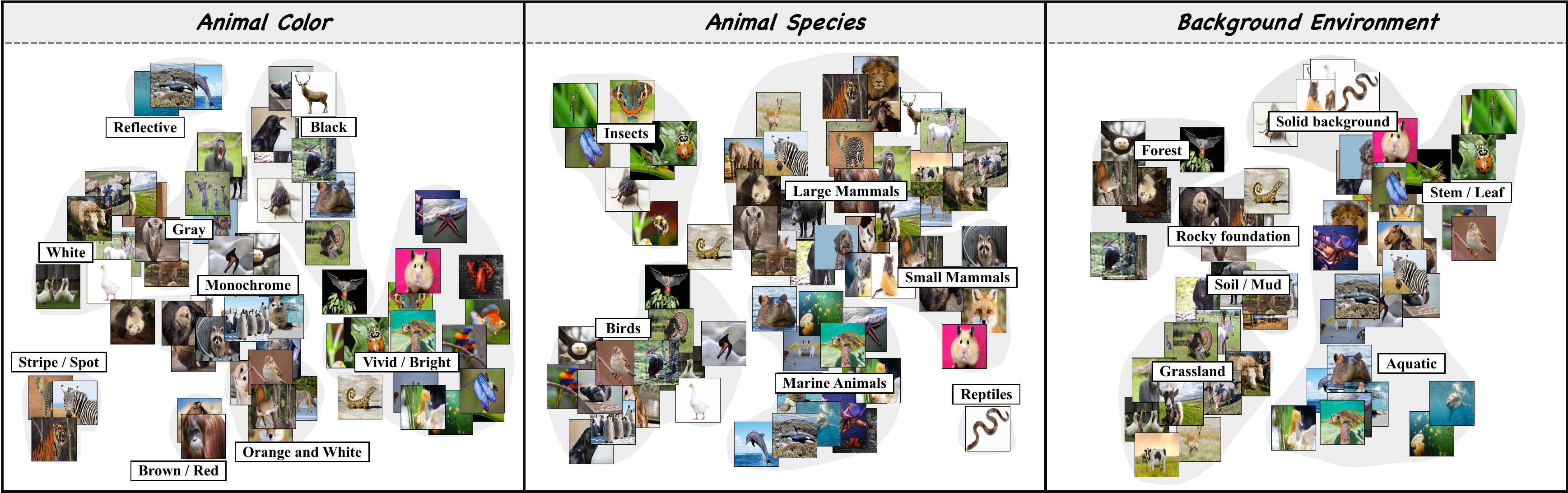}
    \vspace{-3mm}
    \mycaption{T-SNE visualizations of attribute embedding spaces}{
         We visualize the embedding spaces of the same 60 animal images across three different attributes and show that this same set of images is distributed differently and meaningfully across varying attributes.
    }
    \label{fig:tsne_visualization}
\end{figure*}
\begin{figure*}[!ht]
    \centering
    \vspace{+3mm}
    \includegraphics[width=\linewidth]{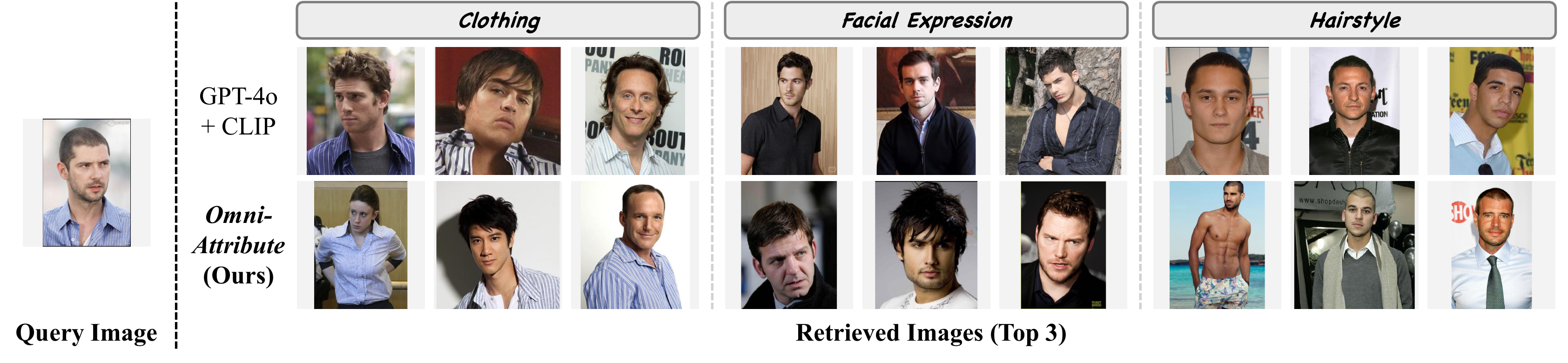}
    \vspace{-5mm}
    \mycaption{Qualitative results of attribute-oriented image retrieval on CelebA~\cite{celeba}}{
        Our embeddings enable image retrieval based on a specified attribute. \ourmodel surpasses the performance of text-guided retrieval by GPT-4o~\cite{gpt_4o} and CLIP~\cite{clip}.
    }
    \label{fig:attribute_retrieval}
\end{figure*}

\subsection{Analysis of Attribute Embeddings}
\label{sec:experiments_analysis}

To gain a deeper understanding of the learned attribute representations, this section qualitatively analyzes them using two methods to enhance interpretability.

\myparagraph{T-SNE Visualizations of Embedding Spaces.}
We sample 60 images from Animal Dataset~\cite{animal_dataset}. For each image, we extract embeddings with three distinct attributes: ``\textit{animal color}'', ``\textit{animal species}'', and ``\textit{background environment}''. We then project them into two dimensions using t-SNE~\cite{tsne} and visualize the attribute-specific embedding spaces in~\cref{fig:tsne_visualization}. Although derived from the same set of images, the embeddings cluster differently and meaningfully depending on the attribute condition, highlighting the model’s ability to disentangle attribute-specific information.

\myparagraph{Attribute-oriented Image Retrieval.}
Next, to demonstrate the model's versatility in the retrieval task, we sample 17.7k images from CelebA~\cite{celeba} and compute embeddings for each image conditioned on three attributes: ``\textit{clothing}'', ``\textit{facial expression},'' and ``\textit{hairstyle}''. Given a single query image, we retrieve the most similar image (measured by the cosine similarity of the pooled embeddings) under each attribute and visualize the results in~\cref{fig:attribute_retrieval}. Since no existing image encoder naively supports this task, we construct a text-guided retrieval baseline by combining GPT-4o~\cite{gpt_4o} and CLIP~\cite{clip} as detailed in~\cref{app:evaluation_retrieval}. Compared to the baseline, \ourmodel captures fine-grained, attribute-specific visual details more accurately, retrieving images with stronger alignment with the target semantic attributes.

\subsection{Ablation Study}
\label{sec:experiments_ablation}
We perform an ablation study to assess the impact of different training strategies and architectural choices. The quantitative results are summarized in~\cref{tab:ablation_study}.

\myparagraph{Cosine Similarity of Attribute Embeddings.}
To evaluate how effectively our embeddings encode attribute-specific information while suppressing unrelated visual concepts, we measure the cosine similarities of two positive attribute embeddings encoded from the semantically linked image pair and apply the same procedure to the negative attribute. High-quality representations are expected to exhibit high similarity for positive attribute embedding pairs and low similarity for negative ones, resulting in a large gap between them. We measure such gaps on 1K validation pairs, each with one randomly selected positive attribute and one negative attribute. The results are reported in~\cref{tab:ablation_study}.

Our observations are as follows:
\begin{itemize}[leftmargin=8pt]
    \item Comparing [a-d] to [e-h], the models without $\mathcal{L}_{\mathrm{con}}$ tend to ignore the attribute conditions and always encode similar embeddings (\ie, trivial $\Delta$), indicating the necessity of contrastive learning for attribute-level representation.
    \item Comparing [c] to [a-b], increasing the number of trainable parameters significantly improves attribute fidelity but slightly degrades text fidelity and image naturalness.
    \item Comparing [d] to [c], full finetuning the MLLM, instead, leads to knowledge forgetting~\cite{lora_vs_full}, resulting in degraded performance on the evaluation set.
    \item From [e-h], hyperparameter choices for contrastive loss play a critical role in balancing high-fidelity encoding and attribute disentanglement. Larger $\lambda_{\mathrm{con}}$ or $\tau$ values ([e], [g]) learn more discriminative attribute embeddings (\ie, larger $\Delta$), but reduce attribute fidelity, and vice versa.
\end{itemize}

With careful designs of the architecture, training strategy, and hyperparameters of $\mathcal{L}_{\mathrm{con}}$, \ourmodel ([i]) achieves the most balanced performance on attribute personalization (\ie, highest average score).
\section{Limitations}
\label{app:limitations}

We identify the following limitations of our work:

\myparagraph{Attribute-specific Embeddings.}
Our attribute embeddings are designed to capture image information related to one or a few specific attributes. This inherently constrains the applicability to tasks such as image editing, where most of the visual content must remain unchanged, and only a limited set of attributes should be modified.

\myparagraph{Entanglement of Correlated Attributes.}
We observe that the model occasionally struggles to disentangle attributes that are often correlated, such as \textit{person identity} and \textit{hairstyle}. For example, as illustrated in~\cref{fig:teaser}, while we attempt to transfer the \textit{identity} of \textit{Vincent van Gogh} to new contexts, the generated images mostly preserve his hairstyle, indicating information leakage. One potential solution is to increase the sampling weight of the \textit{hairstyle dataset} (as depicted in \cref{fig:datasets}(d)) to better learn how to separate these factors. However, it remains an open question whether certain attributes can ever be perfectly disentangled (\eg, whether \textit{hairstyle} is inherently part of \textit{identity}).

\myparagraph{Sensitivity to Contrastive Learning Hyperparameters.}
Prior contrastive learning studies~\cite{simclr,ifnd} noted that the hyperparameters of contrastive loss, such as temperature, typically have a strong and dataset-dependent impact on model performance. In this work, we also notice that the selection of these hyperparameters has a huge impact on the quality of the learned attribute embeddings, as shown in~\cref{tab:ablation_study}.

We leave the study of these limitations for future work.
\section{Conclusion}

We have presented \ourmodel, an open-vocabulary attribute encoder that learns high-fidelity, attribute-specific representations for visual concept personalization. It is achieved through a novel attribute-level annotation strategy with a new dual-objective training framework that jointly optimizes generative fidelity and contrastive disentanglement. The experiments demonstrate the model’s versatility across attribute-based retrieval, personalization, and compositional generation, and show consistent improvements over existing baselines. We believe \ourmodel provides a step toward controllable, interpretable visual representation learning, bridging the gap between multimodal understanding and generative manipulation.

\myparagraph{Impact Statement.}
This work focuses on learning attribute-level image representations for visual concept personalization. The resulting embeddings can facilitate beneficial applications, such as boosting creativity and supporting educational content creation. Beyond these considerations, we do not identify any additional ethical or societal implications beyond those already known to accompany personalized generative modeling.

{
    \small
    \bibliographystyle{ieeenat_fullname}
    \bibliography{main}
}

\appendix
\maketitlesupplementary

\section{Implementation Details}
\label{app:implementation}

\subsection{Training Datasets}
\label{app:implementation_dataset}

\begin{figure}[t]
    \centering
    \includegraphics[trim={0 5mm 0 5mm}, width=\linewidth]{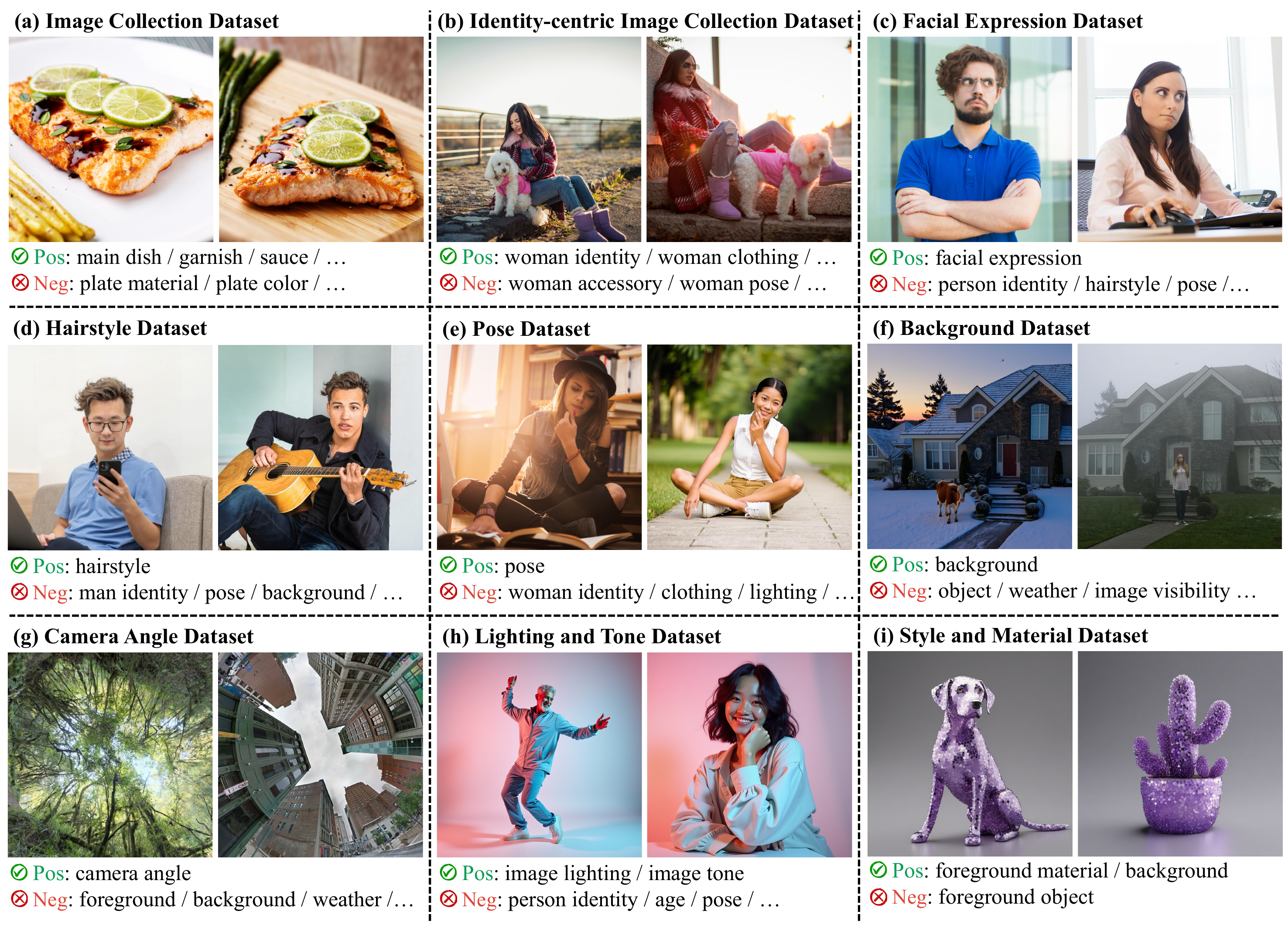}
    \mycaption{Training datasets}{
        Our training image pairs are constructed from two sources: $(i)$ \textit{image collection datasets} (a–b), with image pairs captured during the same photo session, exhibiting varying positive and negative attributes; and $(ii)$ \textit{attribute-specific datasets} (c–i), with image pairs synthesized via generative or editing models that differ mainly in one or a few positive attributes.
    }
    \label{fig:datasets}
\end{figure}

To learn high-quality representations for open-vocabulary attributes, we collect semantically linked image pairs from two complementary sources, resulting in nine datasets, as illustrated in~\cref{fig:datasets}.

First, we collect 23.7M image pairs from an in-house image collection dataset, where images are organized into thematic collections. As shown in~\cref{fig:datasets}(a), images within the same collection are typically captured during a single photo session, exhibiting both shared and distinct characteristics across multiple visual aspects. We randomly sample two images from each theme to form pairs, producing diverse combinations of positive and negative attributes. In total, the dataset contains 600K unique attribute labels. We qualitatively demonstrate the richness and diversity of these labels through a word cloud visualization in~\cref{fig:data_annotation}. To further enhance representations for ``\textit{person identity},'' we additionally sample an identity-centric subset consisting of 2.21M image pairs, where paired images depict the same individual(s), as shown in~\cref{fig:datasets}(b).

While these \textit{image collection datasets} are large-scale and rich in attribute diversity, their image pairs often exhibit multiple entangled positive attributes, making it challenging to isolate one attribute. To address this limitation, we construct seven additional datasets, each focusing on a specific attribute (\eg, \textit{facial expression}, \textit{background}, or \textit{lighting}). As illustrated in~\cref{fig:datasets}(c-i), image pairs in these datasets are designed to share only one or a few positive attributes, facilitating the learning of attribute-specific representations. The detailed curation process for these datasets is summarized as follows:
\begin{table}[h]
    \centering
    \begingroup
    \renewcommand{\arraystretch}{1.4}
    \newcolumntype{L}[1]{>{\raggedright\arraybackslash}m{#1}}
    \newcolumntype{M}[1]{>{\raggedright\arraybackslash}m{#1}}
    \newcolumntype{R}[1]{>{\raggedright\arraybackslash}m{#1}}
    \small
    \label{tab:datasets}
    \vspace{-3mm}
    \begin{tabular}{L{0.14\linewidth} M{0.05\linewidth} R{0.73\linewidth}}
    \toprule
    Dataset & Scale & Curation Process \\
    \midrule
    Facial Expression   & 51.0K & We employ the expression editing model, LivePortrait~\cite{liveportrait}, to manipulate two human images initially with \textit{neutral facial expressions} (according to the detailed descriptions labeled during the attribute annotation). For each pair, we randomly sample a set of editing parameters to generate two images exhibiting the same facial expression. \\
    Hairstyle           & 8.77K & We use an in-house hairstyle editing model to process two face-centric images. Given a reference image with a target hairstyle, the model generates two edited images sharing the same hairstyle. \\
    Pose                & 106K & We extract face, body, and hand keypoints from a human image, and synthesize its paired image using ControlNet~\cite{controlnet} to ensure the same pose as the source image. \\
    Background          & 35.1K & We first sample clean background images filtered by Qwen-VL~\cite{qwenvl2} (by asking whether images depict clear backgrounds). Then, we use Qwen-Image-Edit~\cite{qwen_image_edit} to randomly insert foreground objects and modify contextual factors (\eg, time of day, weather, and camera angle). Each background image is edited twice to form a pair with consistent background. \\
    Camera Angle        & 98.7K & We collect 2,081 panoramic images from multiple sources~\cite{360sod,cvrg_pano,f360isod,saliency_in_VR} and apply PreciseCam~\cite{precisecam} to crop views corresponding to specific camera angles. For each pair, cropping hyperparameters are randomly sampled to ensure both images share the same camera angle. \\
    Lighting and Tone   & 159K & We construct a set of detailed prompts describing various image lighting conditions, along with a separate set of prompts specifying identities and actions. By fixing the lighting prompts while varying the identity prompts, we synthesize paired images via FLUX~\cite{flux}, ensuring consistent lighting across each pair. \\
    Style and Material  & 27.5K & Similar to the \textit{Lighting and Tone} dataset, we design descriptive prompts focusing on style and material properties. Images are synthesized using Stable Diffusion XL~\cite{sdxl}, where the same style or material description is preserved across pairs. \\
    \bottomrule
    \end{tabular}
    \vspace{-3mm}
    \endgroup
\end{table}

During training, we assign a sampling weight of 100 to both image collection datasets, and a weight of 1 to each attribute-specific dataset.

\subsection{Annotation of Positive and Negative Attributes}
\label{app:implementation_annotation}
\begin{figure}[t]
    \centering
    \includegraphics[width=\linewidth]{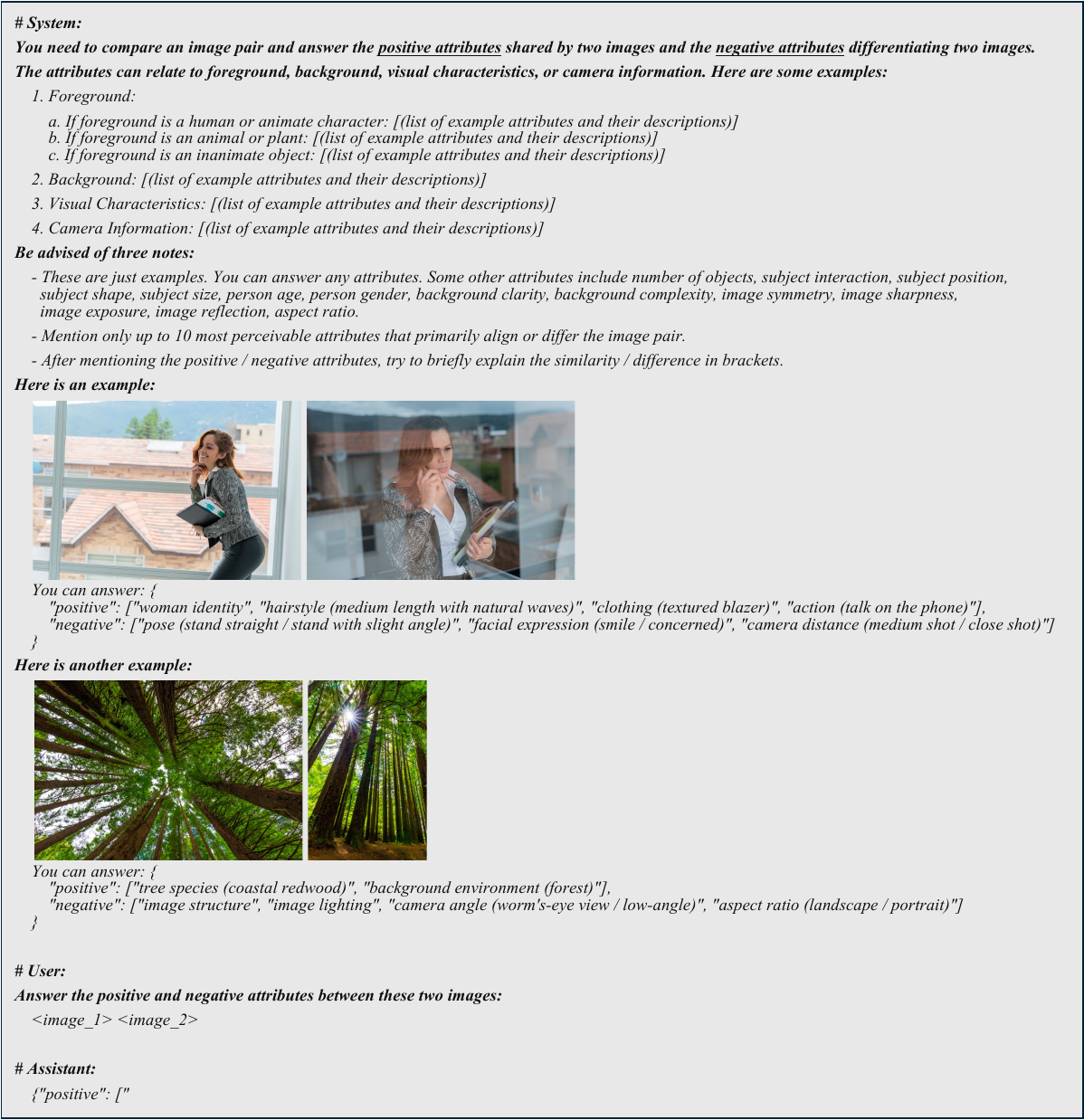}
    \mycaption{Instruction prompt for the first stage of attribute annotation}{}
    \label{fig:instruction_prompt}
\end{figure}

As described in~\cref{sec:methodology_annotation}, we adopt a two-stage annotation pipeline to balance annotation quality and computational cost.

In the first stage, we employ a powerful vision-language model, Qwen2.5-VL-72B~\cite{qwenvl2}, to generate high-quality attribute annotations based on a comprehensive instruction prompt (see \cref{fig:instruction_prompt}). Due to the substantial computational overhead, this stage is applied only to a subset of 200K samples. We introduce two key design choices to further improve annotation effectiveness: $(i)$ Inspired by Chain-of-Thoughts~\cite{chain_of_thought}, we prompt the model to explicitly reason the detailed similarities and differences behind each positive and negative attribute. Empirically, this approach enhances annotation quality. $(ii)$ We design the instruction prompt, ending with the initial segment of the target output format (\ie., ``\texttt{\{positive:[}'' at the bottom of~\cref{fig:instruction_prompt}) and activate the ``\texttt{continue\_final\_message}'' setting. This guides the model to continue generation in the intended structured format and can improve the rate of syntactically valid outputs.

In the second stage, we finetune a Qwen2.5-VL-32B~\cite{qwenvl2} into a task-dependent model tailored for this annotation task, eliminating the need for a lengthy instruction prompt. During training, the input is a concise multimodal prompt: ``\textit{$<$image\_1$>$ $<$image\_2$>$ What are the positive attributes shared by the two images and the negative attributes that differentiate them? For the similarities and differences, explain the reasons in brackets.}'' The output is a string following a structured dictionary-style format. This design reduces the input token length by $3.1\times$ and lowers the per-sample forward latency by $6.3\times$. The model is finetuned on 200K annotated samples using $32\times$ 80GB H100 GPUs, with a learning rate of 2e-7 (with a linear warm-up and cosine decay strategy), a batch size of 512, and 15 training epochs.

For large-scale inference, we enhance computational efficiency by constraining image dimensions so that the total pixel count does not exceed $1280 \times 28 \times 28$. This constraint yields maximum resolutions of $1336 \times 752$ for $16:9$ images, $1157 \times 868$ for $4:3$ images, and $1002 \times 1002$ for square images. We perform inference on 80GB H100 GPUs, where each image pair takes approximately 2.54 seconds to annotate.

\subsection{Model Architecture}
\label{app:implementation_architecture}
As described in~\cref{sec:methodology_architecture}, our attribute encoder consists of a LoRA-finetuned multimodal large language model (MLLM) followed by a fully trainable connector module, while our image generator builds upon a frozen image generator equipped with trainable IP-adapter modules. We detail the model architecture as follows:
\begin{table}[h]
    \centering
    \begingroup
    \renewcommand{\arraystretch}{1.4}
    \newcolumntype{L}[1]{>{\raggedright\arraybackslash}m{#1}}
    \newcolumntype{R}[1]{>{\raggedright\arraybackslash}m{#1}}
    \small
    \label{tab:model_architecture}
    \vspace{-2mm}
    \begin{tabular}{L{0.13\linewidth} R{0.82\linewidth}}
    \toprule
    Module & Description \\
    \midrule
    MLLM            & We adopt Qwen2.5-VL-7B~\cite{qwenvl2} as the base MLLM, where all model parameters are frozen, and LoRA adapters~\cite{lora} are inserted into every linear projection within both the vision encoder and the MLLM modules. Each LoRA module uses a rank of 256 and an alpha value of 512. The token dimensionality is 3584, and we restrict the maximum number of image tokens at 1000. \\
    Connector       & To bridge the MLLM with the image generator, we follow the Step1X-Edit~\cite{step1x_edit} design and incorporate a linear projection layer followed by eight self-attention layers as the connector. The linear layer projects the token dimensionality from 3584 to 4096 to align with the image generator. \\
    Image Generator & We use FLUX.1-dev~\cite{flux} as the base model. Following the observations of Goyal~\etal~\cite{glora}, we note that finetuning the model can diminish FLUX's ability to perform distillation guidance (\ie, it tends to overfit to the non-distillation-guidance setting used during training). To mitigate this, we adopt their proposed \textit{Shortcut-Rerouted Adapter}, which helps preserve the model’s prior for distillation guidance. \\
    IP-Adapter      & We follow the implementation of the InstantX team~\cite{instantx}, where each DiT block~\cite{dit} in FLUX includes two MLP modules (\ie, \texttt{to\_k} and \texttt{to\_v}) that compute key and value embeddings, respectively. These embeddings are then injected into query image tokens through multi-head cross-attention, enabling the generator to incorporate conditioning signals for attribute personalization. \\
    \bottomrule
    \end{tabular}
    \vspace{-1mm}
    \endgroup
\end{table}

\subsection{Model Training}
\label{app:implementation_training}
We train the model in two stages to enhance training efficiency. In the first stage, the model is optimized solely with the generative loss for 100K steps; in the second stage, we introduce an additional contrastive loss and continue training for 10K more steps. This two-stage design is due to the computational overhead of the contrastive loss, which requires four additional forward passes through the MLLM for each training sample (\ie, two images cross-paired with the positive and negative attributes). Therefore, it could substantially slow down convergence if we optimize the contrastive loss from the start.

We conduct all experiments using $64\times$ 80 GB H100 GPUs with a total batch size of 256. Training is performed in mixed precision with parameter and reduction data types set to \texttt{bf16} and \texttt{fp32}, respectively. We apply gradient clipping with a maximum gradient norm of 1.0, and employ Distributed Data Parallel (DDP) and Fully Sharded Data Parallel (FSDP)~\cite{fsdp} strategies for efficient large-scale training. We use the AdamW optimizer~\cite{adamw} with a learning rate of 1e-5, weight decay of 0.01, and $\beta$ of [0.9, 0.99]. A linear warmup is applied during the first 1K steps of both stages. During the first stage, we only finetune the connector and IP-Adapter modules for the first 10K steps while keeping the MLLM parameters frozen to prevent disruption of pretrained representations.

For image preprocessing, we resize each reference image such that its total pixel count does not exceed $1000 \times 28 \times 28$. To improve robustness to low-resolution inputs during inference, we apply a $10\%$ probability of downsampling augmentation. The target images are resized and center-cropped to $512 \times 512$ resolution. For the generative loss, we adopt the flow-matching objective \cite{flow_matching} with $\lambda_{\mathrm{gen}}$ of 1. We adjust the balance between generative and contrastive losses through $\lambda_{\mathrm{con}}$, as ablated in \cref{sec:experiments_ablation} and \cref{tab:ablation_study}.

\section{Evaluation Details}
\label{app:evaluation}

\subsection{Open-vocabulary Attribute Personalization}
\label{app:evaluation_personalization}

\cref{sec:experiments_personalization} compares \ourmodel with the existing models for personalization across 15 attributes, grouped into two categories. We list all evaluation attributes below:
\begin{itemize}[leftmargin=8pt]
    \item \textit{Concrete objects}: man identity, woman identity, object identity, clothing, and background.
    \item \textit{Abstract concepts}: hairstyle, facial expression, makeup, pose, foreground material, texture, camera angle, image lighting, image tone, and artistic style.
\end{itemize}

\begin{table}[t]
    \centering
    \fontsize{8.9pt}{10.5pt}\selectfont
    \begingroup
    \renewcommand{\arraystretch}{1.4}
    \newcolumntype{L}[1]{>{\raggedright\arraybackslash}m{#1}}
    \newcolumntype{R}[1]{>{\raggedright\arraybackslash}m{#1}}
    \mycaption{Full evaluation prompts used in~\cref{fig:qualitative_comparison}}{}
    \vspace{-1mm}
    \label{tab:qualitative_prompts}
    \begin{tabular}{L{0.15\linewidth} R{0.8\linewidth}}
    \toprule
    Reference Attribute & Full Prompt \\
    \midrule
    \textit{Person Identity}     & ``\textit{A woman in a side view that looks festive and blowing a party horn with a laughing expression, surrounded by colorful streamers and balloons.}'' \\
    \textit{Dog Identity}        & ``\textit{A dog playing in a bright, minimalist art gallery with polished floors and white walls displaying modern paintings. Natural light filters through glass doors, highlighting a clean, contemporary aesthetic of calm and creativity.}'' \\
    \textit{Clothing}            & ``\textit{A joyful young woman with long brunette hair leaping midair in a white space, frozen in motion. Her athletic form radiates freedom, energy, and the graceful strength of dance.}'' \\
    \textit{Hairstyle}           & ``\textit{A woman wearing a denim jacket over a white top. Her ears are adorned with large hoop earrings, and she has bold makeup featuring defined eyebrows and bright lipstick. The background is a soft, solid pink hue.}'' \\
    \textit{Facial Expression}   & ``\textit{A woman with dark hair styled in a messy bun. She is wearing a plain white shirt, and the background is a clean, neutral white.}'' \\
    \textit{Makeup}              & ``\textit{A woman with long, wavy brown hair, wearing a white t-shirt featuring bold red and blue graphics. The background includes a bright turquoise frame against a neutral-colored wall, creating a striking contrast.}'' \\
    \textit{Pose}                & ``\textit{A low-angle view a joyful young woman midair against a bright blue sky. Wearing orange trousers and green sneakers, she embodies freedom, exuberance, and carefree spontaneity.}'' \\
    \textit{Foreground Material} & ``\textit{A handbag equipped with a chain strap resting on a clean white block. The background is a soft gradient of light blue, creating a calm and sophisticated setting.}'' \\
    \textit{Texture}             & ``\textit{A sleek Mclaren sports car parked on a paved street. The vehicle's aerodynamic design and shiny exterior reflect its high-performance nature. In the background, a brick building with large windows and various utility fixtures stands, along with a striped awning and some greenery.}'' \\
    \textit{Image Lighting}      & ``\textit{A pristine armchair with a tufted backrest and decorative buttons standing against a clean, minimalist backdrop. Its elegant design features slender, spindle-like legs that add a touch of classic charm to its modern aesthetic. The seat cushion appears soft and inviting, complementing the chair's overall refined look.}'' \\
    \textit{Artistic Style}      & ``\textit{A graceful white horse galloping across a dirt path, its mane and tail flowing in the breeze. The background is a lush, dense forest bathed in soft, golden light, creating a serene and natural setting.}'' \\
    \bottomrule
    \end{tabular}
    \endgroup
\end{table}
\begin{table}[t]
    \centering
    \small
    \mycaption{Numerical results of open-vocabulary attribute personalization}{
        We complement the quantitative comparison graphs (\cref{fig:quantitative_comparison}) by providing the exact measurements of \textit{text fidelity} (Text-F), \textit{attribute fidelity} (Attr-F), \textit{image naturalness} (Natural), and their average.
    }
    \label{tab:quantitative_numerical}
    \begin{tabular}{lcccccccc}
    \toprule
    \multirow{2}{*}{Method} & \multicolumn{4}{c}{\textit{Concrete Objects}} & \multicolumn{4}{c}{\textit{Abstract Concepts}}\\
    \cmidrule(lr){2-5} \cmidrule(lr){6-9}
    & Text-F$\uparrow$ & Attr-F$\uparrow$ & Natural$\uparrow$ & Average$\uparrow$ &
      Text-F$\uparrow$ & Attr-F$\uparrow$ & Natural$\uparrow$ & Average$\uparrow$ \\
    \midrule
    \midrule
    \multicolumn{2}{l}{\textit{MLLM Evaluation}} &&&&&&& \\
        CLIP~\cite{clip} &
        0.9000 & 0.6550 & 0.8400 & 0.7983 &
        0.9504 & 0.3120 & 0.8056 & \cellthird 0.6893 \\
        DINOv2~\cite{dinov2} &
        0.8460 & 0.7747 & 0.8160 & 0.8122 &
        0.9168 & 0.3568 & 0.8073 & \cellsecond 0.6936 \\
        Qwen-VL~\cite{qwenvl2} &
        0.8820 & 0.6848 & 0.8460 & 0.8043 &
        0.9760 & 0.2736 & 0.8072 & 0.6856 \\
        OmniGen2~\cite{omnigen} &
        0.9140 & 0.7890 & 0.7810 & \cellsecond 0.8280 &
        0.9512 & 0.2863 & 0.7435 & 0.6603 \\
        FLUX-Kontext~\cite{flux_kontext} &
        0.8540 & 0.8910 & 0.7091 & \cellthird 0.8180 &
        0.9304 & 0.3363 & 0.7440 & 0.6702 \\
        Qwen-Image-Edit~\cite{qwen_image_edit} &
        0.5910 & 0.8980 & 0.8091 & 0.7660 &
        0.3872 & 0.7515 & 0.6895 & 0.6074 \\
        \midrule
        \textbf{\ourmodel} &
        0.9381 & 0.7634 & 0.8540 & \cellfirst 0.8518 &
        0.8539 & 0.5181 & 0.8079 & \cellfirst 0.7267 \\
    \midrule
    \midrule
    \multicolumn{2}{l}{\textit{Human Evaluation}} &&&&&&& \\
        CLIP~\cite{clip} &
        0.9300 & 0.6050 & 0.8477 & 0.7942 &
        0.9159 & 0.4331 & 0.8576 & 0.7356 \\
        DINOv2~\cite{dinov2} &
        0.9087 & 0.6844 & 0.8434 & 0.8122 &
        0.9179 & 0.4363 & 0.8684 & 0.7409 \\
        Qwen-VL~\cite{qwenvl2} &
        0.9242 & 0.5505 & 0.8542 & 0.7763 &
        0.9445 & 0.3957 & 0.8754 & 0.7385 \\
        OmniGen2~\cite{omnigen} &
        0.9376 & 0.7575 & 0.8110 & \cellthird 0.8354 &
        0.9462 & 0.4392 & 0.8533 & \cellthird 0.7463 \\
        FLUX-Kontext~\cite{flux_kontext} &
        0.9054 & 0.8785 & 0.8032 & \cellsecond 0.8624 &
        0.9323 & 0.4688 & 0.8910 & \cellsecond 0.7641 \\
        Qwen-Image-Edit~\cite{qwen_image_edit} &
        0.7055 & 0.8913 & 0.8426 & 0.8131 &
        0.6133 & 0.8622 & 0.8618 & 0.7344 \\
        \midrule
        \textbf{\ourmodel} &
        0.9564 & 0.7691 & 0.8680 & \cellfirst 0.8645 &
        0.9374 & 0.7031 & 0.9584 & \cellfirst 0.8663 \\
    \bottomrule
    \end{tabular}
\end{table}

To complement the qualitative and quantitative evaluation shown in \cref{fig:qualitative_comparison,fig:quantitative_comparison}, we list the full evaluation prompts used in \cref{fig:qualitative_comparison} in \cref{tab:qualitative_prompts}, and report the original numerical values visualized in \cref{fig:quantitative_comparison} in \cref{tab:quantitative_numerical}.

\begin{figure}[t]
    \centering
    \includegraphics[width=.85\linewidth]{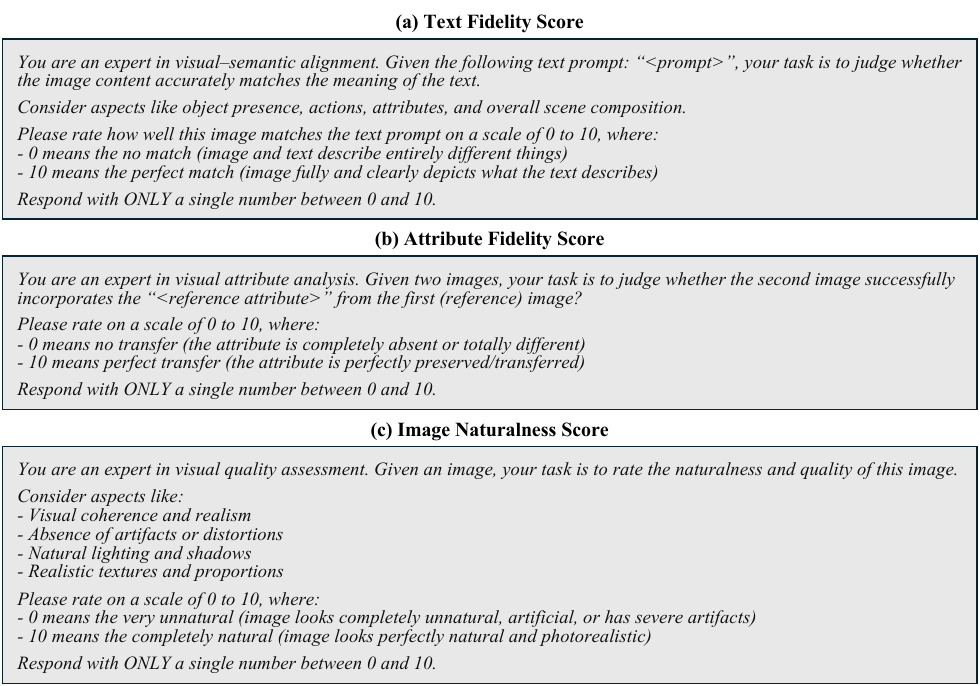}
    \mycaption{Instruction prompt for MLLM evaluation}{}
    \label{fig:mllm_evaluation_prompt}
\end{figure}
\begin{figure}[ht]
    \centering
    \includegraphics[width=\linewidth]{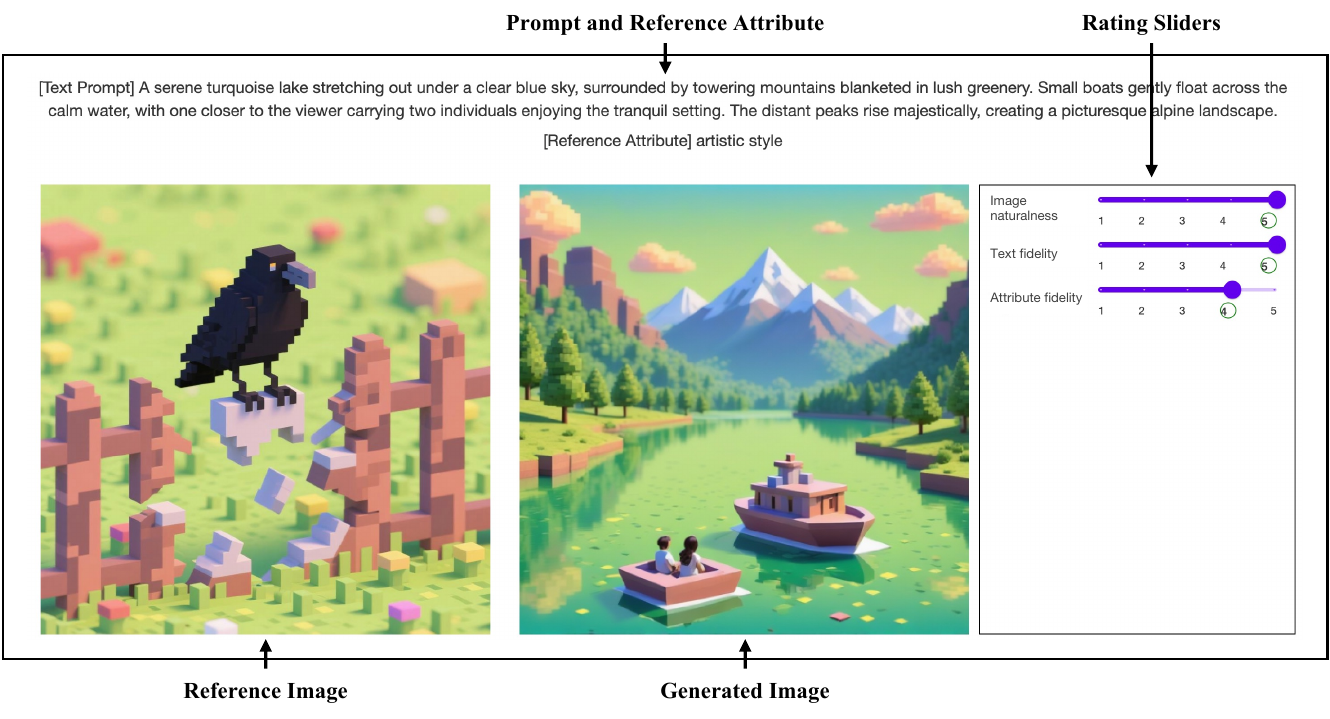}
    \vspace{-7mm}
    \mycaption{Interface of the user study}{
        Given the input conditions (top and right) and the generated image (center), participants are asked to rate three aspects: \textit{image naturalness}, \textit{text fidelity}, and \textit{attribute fidelity} on a 1 (poor) to 5 (excellent) scale using the sliders (left).
    }
    \label{fig:user_study_interface}
\end{figure}

As described in~\cref{sec:experiments_personalization}, we apply both MLLM-based and human evaluations for a comprehensive assessment. For the MLLM evaluation, we query GPT-4o~\cite{gpt_4o} three times using the prompts shown in~\cref{fig:mllm_evaluation_prompt} to measure \textit{text fidelity}, \textit{image fidelity}, and \textit{image naturalness}. For the user study, participants are presented with the reference image-attribute pair, the prompt, and the generated image for each sample, as shown in~\cref{fig:user_study_interface}. They are then asked to rate the three evaluation metrics on a scale from 1 (poor) to 5 (excellent). All scores are subsequently normalized to the range of [0,1].

\clearpage
\subsection{Attribute-oriented Image Retrieval}
\label{app:evaluation_retrieval}
Since there is no existing model directly supporting attribute-oriented image retrieval, we construct a text-guided baseline using GPT-4o~\cite{gpt_4o} and CLIP~\cite{clip}. Specifically, we first prompt GPT-4o to generate descriptive texts of approximately 60 words for each target attribute. These descriptions are then converted into text embeddings using CLIP, which are subsequently used to retrieve the most semantically similar images corresponding to the given attribute.

\section{Additional Results}
\label{app:additional_results}
\begin{figure}[ht]
    \centering
    \includegraphics[width=\linewidth]{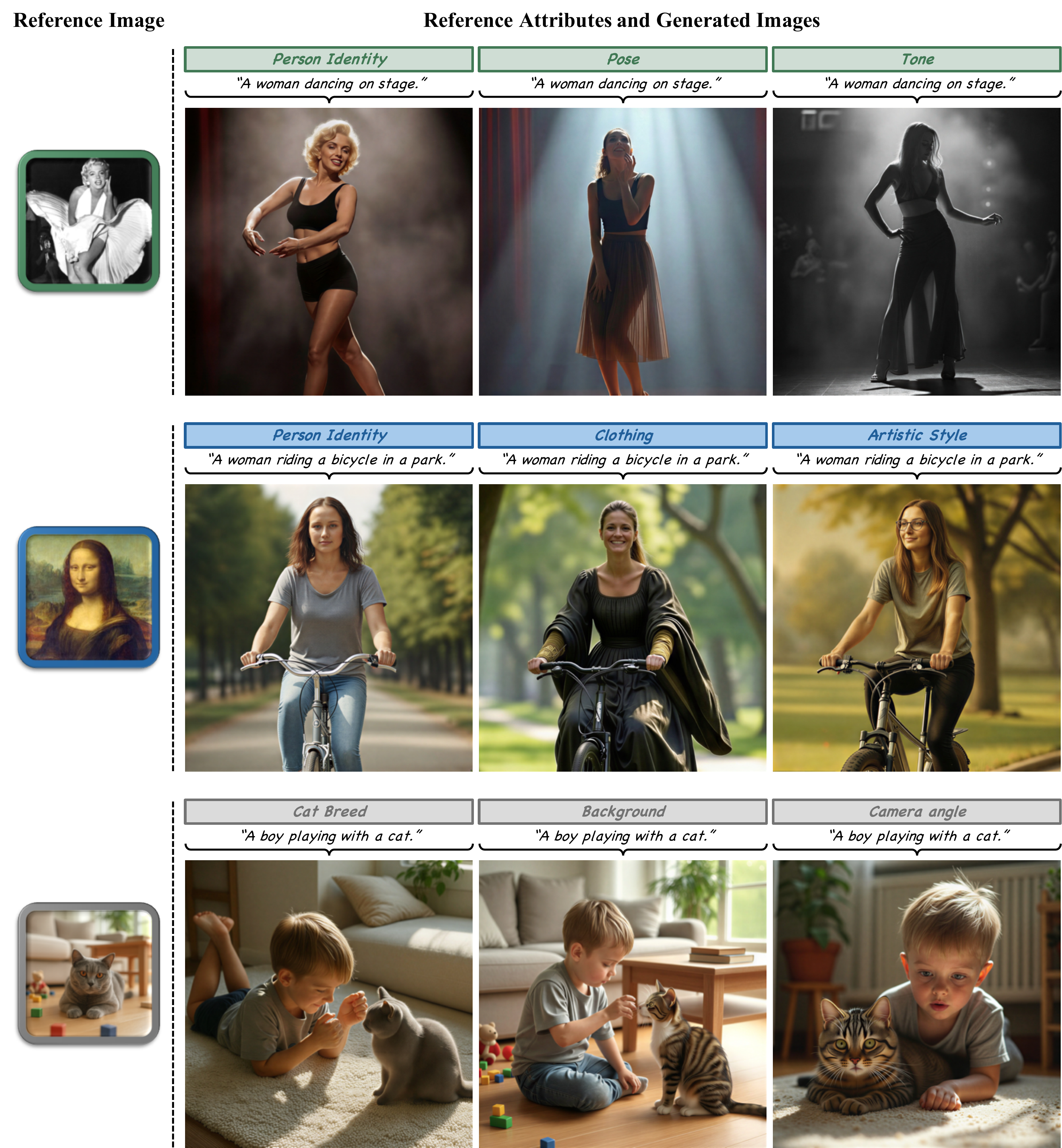}
    \mycaption{Additional results of attribute disentanglement}{
        Each row shows three generated images (right), which are conditioned on the same reference image (left) and the same textual prompt, but with different attribute inputs (colored boxes). As seen, given the same reference image, \ourmodel effectively extracts attribute-specific representations, enabling the coherent synthesis of the user-specified attribute in new contexts while reducing the leakage of irrelevant visual information from the reference image. 
    }
    \label{fig:attribute_disentanglement}
\end{figure}

\cref{fig:teaser}(a) illustrates that \ourmodel can extract high-fidelity, attribute-specific information while suppressing irrelevant visual details. This helps reduce ``\textit{copy-and-paste}'' artifacts and leads to a more coherent synthesis of the user-specified attribute in new contexts. Additional results demonstrating such attribute disentanglement are shown in~\cref{fig:attribute_disentanglement}.

\begin{figure}[ht]
    \centering
    \includegraphics[width=\linewidth]{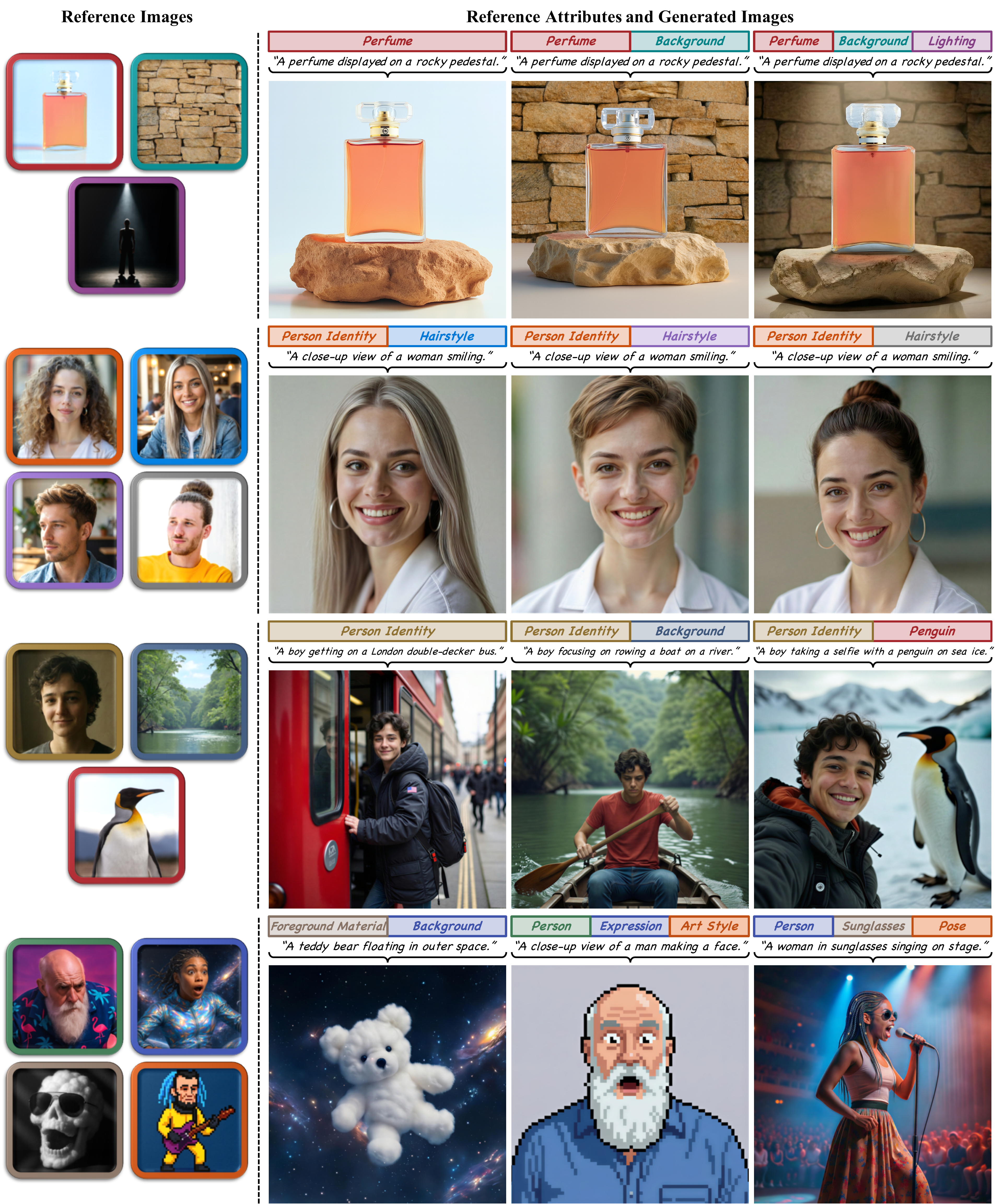}
    \mycaption{Practical and creative applications of \ourmodel}{
        From top to bottom, each row demonstrates the practical utility of \ourmodel across four real-world applications: $(i)$ advertisement image synthesis, $(ii)$ hairstyle customization, $(iii)$ storytelling visualization, and $(iv)$ creative content generation.
    }
    \label{fig:applications}
\end{figure}

To further showcase the practical utility of \ourmodel, we design four real-world application scenarios: $(i)$ advertisement image synthesis, $(ii)$ hairstyle customization, $(iii)$ storytelling visualization, and $(iv)$ creative content generation. The corresponding results are shown in~\cref{fig:applications}.


\end{document}